\documentclass[man,natbib,floatsintext]{apa6}
\usepackage[utf8]{inputenc}
\usepackage{graphicx}
\usepackage{amsmath}
\usepackage{amssymb}
\usepackage{url}
\usepackage{subcaption}
\usepackage[font=small,labelfont=bf]{caption}
\usepackage[usestackEOL]{stackengine}

\usepackage{setspace}
\AtBeginEnvironment{tabular}{\singlespacing}

\title{MARMOT: A Deep Learning Framework for Constructing Multimodal Representations for Vision-and-Language Tasks}
\shorttitle{MARMOT}
\twoauthors{Patrick Y. Wu}{Walter R. Mebane, Jr.}
\twoaffiliations{Department of Political Science, University of Michigan}{Department of Political Science and Department of Statistics, University of Michigan}

\abstract{Political activity on social media presents a data-rich window into political behavior, but the vast amount of data means that almost all content analyses of social media require a data labeling step. However, most automated machine classification methods ignore the multimodality of posted content, focusing either on text or images. State-of-the-art vision-and-language models are unusable for most political science research: they require all observations to have both image and text and require computationally expensive pretraining. This paper proposes a novel vision-and-language framework called \textbf{m}ultimod\textbf{a}l \textbf{r}epresentations using \textbf{mo}dality \textbf{t}ranslation (MARMOT). MARMOT presents two methodological contributions: it can construct representations for observations missing image or text, and it replaces the computationally expensive pretraining with modality translation. MARMOT outperforms an ensemble text-only classifier in 19 of 20 categories in multilabel classifications of tweets reporting election incidents during the 2016 U.S. general election. Moreover, MARMOT shows significant improvements over the results of benchmark multimodal models on the Hateful Memes dataset, improving the best result set by VisualBERT in terms of accuracy from 0.6473 to 0.6760 and area under the receiver operating characteristic curve (AUC) from 0.7141 to 0.7530. The GitHub repository for MARMOT can be found at \url{github.com/patrickywu/MARMOT}.}

\authornote{\noindent This work was supported in part by an NSF RIDIR grant under award number SES-1925693 (``The Sub-National Data Archive System for Social and Behavioral Data'') and a fellowship from the Michigan Institute for Computational Discovery and Engineering (MICDE). We would like to thank Michael Bailey, Pablo Barber\'a, Ben Lempert, Emily Mower Provost, Kevin Quinn, Sarah Shugars, Stuart Soroka, and two anonymous reviewers for helpful suggestions and comments.}

\begin{document}

\maketitle

\section{Introduction}
\label{sec:intro}
This paper introduces a novel deep learning framework for constructing representations for vision-and-language tasks usable for political science and communications research of social media. This framework seeks to improve the classification or labeling step of research on social media content, which usually ignores either the image or the post's text. For example, \citet{Barbera.Casas.Nagler.Egan.Bonneau.Jost.Tucker2019} use latent Dirichlet allocation \citep{blei.ng.jordan.2003} to cluster text of tweets by topic. \citet{Mebane.Wu.Woods.Klaver.Pineda.Miller2018} use active learning with support vector machines and an ensemble classifier to sort the text of tweets into categories and subcategories. \citet{pan_siegel_2020} use a crowdsourcing approach to label the text of tweets into specific categories and sentiment. \citet{casas_williams_2019} also use a crowdsourcing approach to label images from tweets into the emotions they were supposed to invoke. All such examples label, classify, or cluster posts using one modality. 

\begin{figure}[!ht]
    \centering
    \includegraphics[scale=0.20]{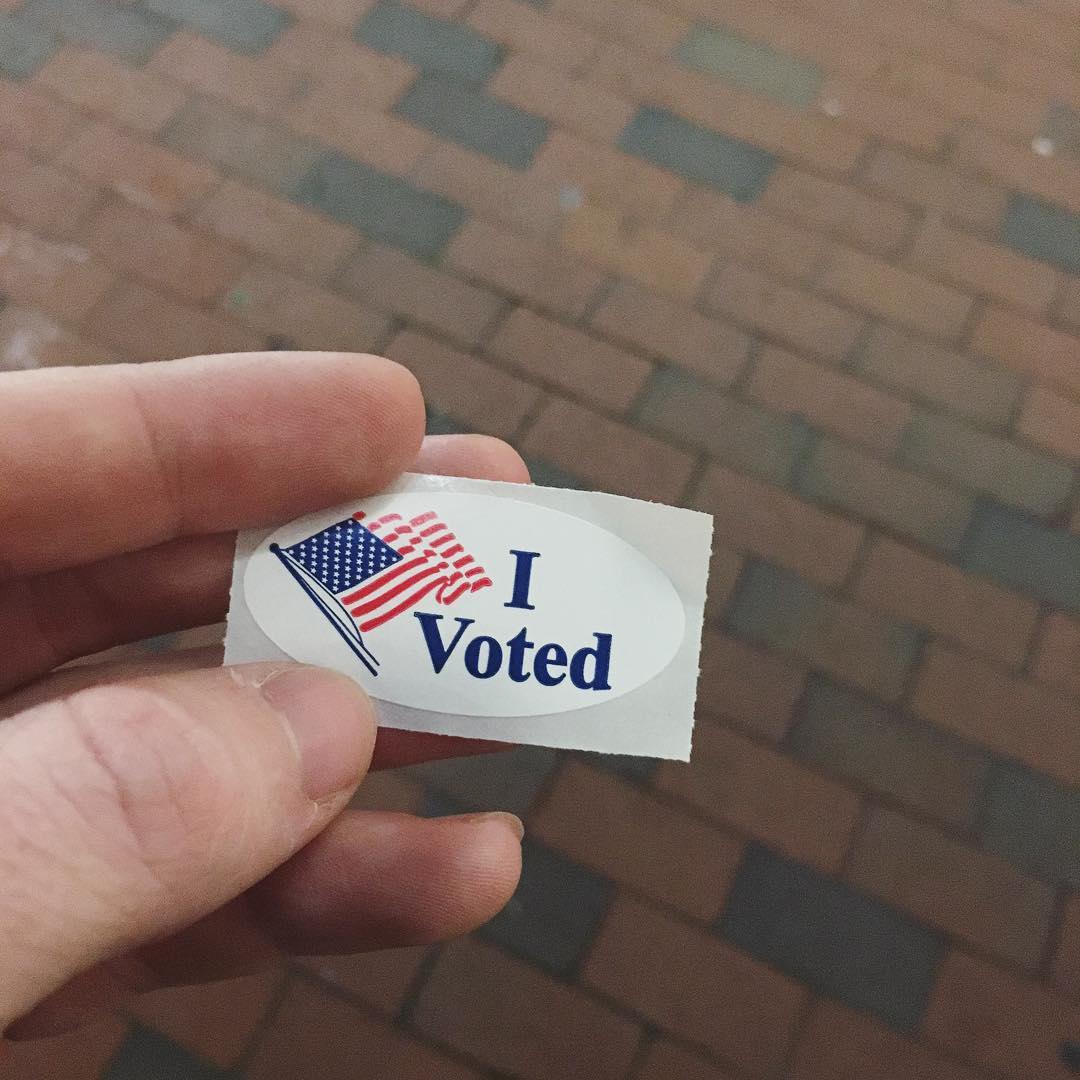}
    \caption{``Done and done \#vote''}
    \label{fig:example1}
\end{figure}

A unimodal focus can potentially create biases in the processed data, leading to misleading inferences in the downstream analysis of the data. Both image and text must be considered to reduce these potential biases. For example, consider the following tweet in Figure \ref{fig:example1}. If we were interested in classifying tweets as reports of being able to vote with no reported problems, this tweet would fit those criteria: the text indicates that the person finished something, along with the vote hashtag, and the image of the ``I Voted'' sticker indicates that the person voted. Thus, the combination of the text and the image indicates that the person could successfully vote, meeting the labeling criteria. However, if another tweet contained a similar image but the text indicated that they encountered problems voting, such a tweet would not fit the labeling criteria. Only by jointly considering the image and text can we definitively conclude that the person in Figure \ref{fig:example1} was able to vote with no reported problems.

An approach to machine classification of multimodal posts would be to jointly map image-text combinations to $d$-dimensional vectors, an approach known as representation learning. Representation learning methods automatically learn the mapping of observations to vector representations. Popular examples of representation learning methods include word2vec \citep{mikolov.sutskever.chen.corrado.dean.2013}, GloVe \citep{pennington2014glove}, and doc2vec \citep{le_mikolov_2014}. 

Representation learning methods exist that produce joint representations of images and text. Some of these models, known as late fusion approaches, use separate models for image and text and then combine the outputs of these two models \citep{liu2018learn}. These models usually allow for missing modalities among observations but fail to effectively learn patterns between images and text. Other models, known as early fusion approaches, input both image and text features into a single model. Early fusion models efficiently learn patterns between the image and text directly \citep{liu2018learn}. Most state-of-the-art approaches use an early fusion approach. 

State-of-the-art approaches such as ViLBERT \citep{lu2019vilbert} and VisualBERT \citep{li2019visualbert} are not well-suited for political science and communications research. Early fusion approaches usually require all observations have image and text, which is almost always not true for social media data that social scientists are interested in---posts may contain text, image(s), or some combination of both. Most state-of-the-art early fusion models, most of which are based on transformers \citep{vaswani.2017}, are also pretrained on image annotation datasets, such as Microsoft COCO \citep{lin2014microsoft}. These image annotation datasets contain images with associated captions. This pretraining serves two purposes. First, it adapts the underlying transformers-based pretrained language model, originally trained to accept text input only, to accept as input both text and image features. Second, it learns the relationship between the text and image. However, because most real-world multimodal observations that need to be classified are not simply a caption describing what is happening in an image, additional pretraining is needed using the data of interest in order for the model to adapt to the target domain. Pretraining is also computationally expensive, requiring computational resources not available to most social scientists.  

To this end, we propose \textbf{m}ultimod\textbf{a}l \textbf{r}epresentations from \textbf{mo}dality \textbf{t}ranslation, or MARMOT, a novel transformers-based architecture that produces multimodal representations. MARMOT aims to solve the two issues that make state-of-the-art multimodal models unusable for political science and communications research. First, we use attention masks to handle missing modalities explicitly. Attention masks, typically used in machine translation and question-answering tasks, are used to prevent the self-attention mechanism of the transformer from attending to missing modalities, meaning that representations can be constructed even for observations missing modalities. Second, to capture the spirit of pretraining while avoiding expensive computational costs, we propose modality translation. Instead of pretraining our model on an image annotation dataset, we directly generate captions for each observation containing an image using a pretrained image captioner. To avoid having to adapt the underlying transformers-based pretrained language model to accept both text and image features, we use a transformer decoder initialized with pretrained BERT weights \citep{devlin.chang.lee.toutanova.2018}. The image captions, derived from a pretrained image captioner such as self-critical sequence training \citep{rennie2016selfcritical}, are inputted into the BERT decoder, and the image features, derived from a pretrained image network such as ResNet-152 \citep{he2015deep}, are inputted at the encoder-decoder attention layer. The BERT decoder constructs what we call the translated image. Modality translation maps image features to the relevant parts of the text feature space. The last step jointly inputs the text, image captions, and translated image features into a transformer encoder initialized with pretrained BERT weights. The output of the transformer encoder is the joint image-text representation. 

MARMOT makes two methodological contributions. First, it introduces modality translation. Modality translation replaces the computationally expensive pretraining process and allows the model to learn directly from the data of interest. Second, the model can calculate representations even for observations missing an image or text. 

We apply MARMOT to data classification tasks on two datasets. The first is a dataset of tweets reporting election incidents during the 2016 U.S. general election \citep{Mebane.Wu.Woods.Klaver.Pineda.Miller2018}. All tweets contain text but some tweets contain an image. MARMOT outperforms the text-only classifier used in \citet{Mebane.Wu.Woods.Klaver.Pineda.Miller2018} on 19 of 20 categories in multilabel classifications of tweets (and equals the performance in the last category). The second is the Hateful Memes dataset, recently released by Facebook Research to test multimodal models \citep{kiela2020hateful}. The goal is to classify each meme as hateful or not. Detecting hateful speech and memes is of interest to both computer science \citep[e.g.,][]{davidson2017automated, macavaneyetal} and political science \citep[e.g.,][]{trumphateontwitter, siegel_badaan_2020}. This dataset also contains text and an image for each observation, making MARMOT comparable to other state-of-the-art multimodal models. MARMOT improves upon the results set by benchmark state-of-the-art multimodal models on this dataset, even outperforming pretrained multimodal models. It improves the best benchmark result set by VisualBERT \citep{li2019visualbert} pretrained on MS COCO in terms of accuracy from 0.6473 to 0.6760 and in terms of area under the receiver operating characteristic curve (AUC) from 0.7141 to 0.7530. The model finished in the top 1\% of all participants in the Hateful Memes challenge. 

The paper proceeds as follows. First, we review the literature on multimodal models. We then detail the architecture of MARMOT. We apply MARMOT to a dataset of election incidents reported on Twitter during the 2016 U.S. general election and the Hateful Memes dataset. We then make concluding remarks and discuss future directions of the project.

\section{Approaches to Classifying Multimodal Data}
A \textit{modality} is defined as some item that provides information to a reader or viewer, such as text, audio, images, or video. It is also common for multiple modalities to exist together, providing readers with the task of jointly considering the modalities to understand the information conveyed \citep{Mogadala_polylingualmultimodal}. Our paper considers image and text, but other works have focused on other combinations of modalities as well, such as text and video \citep[see, e.g.,][]{sun2019videobert}. 

The goal is to develop a model usable for social scientists labeling data that consist of both images and text. That means that the model must work with small datasets, it must be able to run with modest computational resources, and it must handle data that may have observations missing modalities. We look at the previous models and works that we leverage within MARMOT to develop a model that satisfies these requirements. 

\subsection{Late Fusion vs. Early Fusion Models}
Early works on multimodal learning largely focused on late fusion approaches, a set of multimodal learning models where individual modalities are inputted into separate models. The outputs of these separate models are then combined through a policy. Initial works used major features of the images and bag-of-words approaches \citep{tian.zheng.zhu.2012}. Later works used deep learning methods such as deep convolutional neural networks \citep{zahavy2016picture}. Social science methodologists have also developed late fusion approaches. \citet{zhang.pan.2019} use a late fusion approach with a convolutional neural network for images and a recurrent neural network for text in order to identify Weibo posts that discuss offline collective action. The main advantage of late fusion approaches is that observations can be missing modalities: one could use the representation generated by the text model alone, the representation from the image model alone, or the combined representation from the two models. Nevertheless, because there are separate models for each modality, such an approach cannot learn interactions or patterns across the modalities in a meaningful fashion. 

Early fusion approaches, on the other hand, create joint representations of images and text. A single model is used to learn within and across both modalities, which is its key advantage. It assumes, however, that one model is suitable for both modalities. \citet{wang2019makes} note that early fusion multimodal networks often perform poorly because using a single optimization strategy is almost always suboptimal for a model that deals with multiple modalities. 

Notwithstanding these issues, early fusion approaches have quickly risen in popularity with the development of the transformer and pretrained language models such as BERT. Some of these self-supervised architectures include VisualBERT \citep{li2019visualbert}, Visual-Linguistic BERT \citep[VL-BERT,][]{su2019vlbert}, Vision and Language BERT \citep[ViLBERT,][]{lu2019vilbert}, Learning Cross-Modality Encoder Representations from Transformers \cite[LXMERT,][]{tan2019lxmert}, and Multimodal Bitransformers \citep[MMBT,][]{kiela2019supervised}. 

\subsection{Attention and Transformers}
The transformer consists of four components: the attention mechanism, layer normalization, residual connections, and the feedforward layer. To support understanding the transformers that MARMOT is based on and the attention mask feature of MARMOT, we review the attention mechanism and transformers here; brief introductions to layer normalization, residual connections, and the feedforward layer can be found in the Supplemental Information. More technical introductions about attention and transformers can be found at \citet{rush_2018} and \citet{bloem_2019}. 

\subsubsection{Attention}
\label{sec:about_attention}
Transformers use attention \citep{bahdanau2014neural} in the self-attention layer and the encoder-decoder attention layer. Rather than directly defining attention, we first turn to the more intuitive self-attention. Attention is a generalization of self-attention. 

Self-attention is a sequence-to-sequence operation, meaning that a sequence of vectors is inputted and a sequence of vectors is outputted. Self-attention relates all positions of a sequence with one another in order to compute a new representation of the same sequence. To make this idea more concrete, we start with a simplified version of self-attention. Denote the input vectors as $\mathbf{x}_1, \mathbf{x}_2, ...., \mathbf{x}_N$ and denote the output vectors, the new representation of the sequence of vectors $\mathbf{x}$, as $\mathbf{z}_1, \mathbf{z}_2, ..., \mathbf{z}_N$. We can assume that all vectors $\mathbf{x}$ and $\mathbf{z}$ have dimension $k$. To calculate $\mathbf{z}_i$, simplified self-attention simply takes a weighted sum over all input vectors $\mathbf{x}_j$, for $j \in \{1,...,N\}$: 
\begin{equation}
    \mathbf{z}_i = \sum_j w_{ij} \mathbf{x}_j
    \label{eq:weightedsimple}
\end{equation}
The weight $w_{ij}$ is not a learned parameter, but is calculated from a similarity function over $\mathbf{x}_i$ and $\mathbf{x}_j$, such as the dot product: $w_{ij}' = \mathbf{x}_i^T \mathbf{x}_j$. Because the dot product calculates a weight that is between negative and positive infinity, we apply a softmax function to map the weights $w_{ij}'$ to $[0,1]$ and to ensure they sum to 1 over the entire sequence: 
\begin{equation}
    w_{ij} = \frac{\exp(w_{ij}')}{\sum_{l} \exp(w_{il}')}
    \label{eq:softmaxattention}
\end{equation}
$w_{ij}$ are known as the attention weights because they indicate how much attention the $i$th output vector $\mathbf{z}_i$ should pay to the $j$th input vector $\mathbf{x}_j$. The core goal of self-attention is propagating information between input vectors. 

There are a few more modifications to give self-attention more representational power. Notice that each input vector $\mathbf{x}_i$ is used in three ways: (1) it is compared to every other input vector $\mathbf{x}_j$, $j \in \{1,...,N\}$, to calculate the weights for output $\mathbf{z}_i$; (2) it is compared to every other input vector to calculate the weights for all other outputs $\mathbf{z}_j$, $j \in \{1,...,i-1,i+1,...,N\}$; (3) and it is used as a part of the weighted sum in equation \ref{eq:weightedsimple} to compute each output vector. These roles are called the query, the key, and the value, respectively. To allow the query, key, and value to differ, we can define them as $\mathbf{q}_i = \mathbf{W}_q \mathbf{x}_i$, $\mathbf{k}_i = \mathbf{W}_k \mathbf{x}_i$ and $\mathbf{v}_i = \mathbf{W}_v \mathbf{x}_i$, respectively, where $\mathbf{W}_q$, $\mathbf{W}_k$, and $\mathbf{W}_v$ are $k \times k$ weight matrices. To give even more representational power, we can use $h$ query, key, and value weight matrices, yielding $h$ separate sets of output vectors that can be combined via a linear transformation. This is known as multiheaded attention. 

The dot product is rarely used as the similarity function to calculate the weights because the softmax function is sensitive to large input values, affecting gradients in backpropagation. Because the average value of the dot product grows with $k$, we divide the dot product by $\sqrt{k}$, called the scaled dot product. 

The following three equations reflect the above discussion and define self-attention: 
\begin{equation}
    w_{ij}' = \mathbf{q}_i^T \mathbf{k}_j
    \label{eq:weight_unsoftmaxed}
\end{equation}
\begin{equation}
    w_{ij} = \text{softmax}(w_{ij}')
    \label{eq:weight_softmax}
\end{equation}
\begin{equation}
    \mathbf{z}_i = \sum_j w_{ij} \mathbf{v}_j
    \label{eq:weightsfull}
\end{equation}

Notice that self-attention is permutation invariant: nothing about Equations \ref{eq:weight_unsoftmaxed} to \ref{eq:weightsfull} takes into account the order of the vectors in the sequence $\mathbf{x}$. The attention weights derived for the sentence ``Trump beat Clinton in the 2016 U.S. general election'' would be the same as the attention weights derived for the sentence ``Clinton beat Trump in the 2016 U.S. general election,'' even though the word order reveals a key distinction in the two sentences. We use positional embeddings to make the two sentences distinct. Positional embeddings map the word position to either learnable embeddings or fixed embeddings. These positional embeddings are then added to the input. Absolute position embeddings, where a unique embedding is learned for each position, are the most popular positional embedding choice. Fixed sinusoidal positional embeddings also work well in many contexts \citep{vaswani.2017}.  

Generalized attention resembles self-attention in every manner except that the queries, keys, and values are not all based on the same sequence of vectors $\mathbf{x}$. The encoder-decoder attention layer defines the query as $\mathbf{q}_i = \mathbf{W}_{q} \mathbf{x}_i$, the key as $\mathbf{k}_i = \mathbf{W}_k \mathbf{y}_i$, and the value as $\mathbf{v}_i = \mathbf{W}_v \mathbf{y}_i$, where $\mathbf{x}_i$ comes from a sequence of vectors $\mathbf{x}$ and $\mathbf{y}_i$ comes from a separate sequence of vectors $\mathbf{y}$. Other than this, Equations \ref{eq:weight_unsoftmaxed}, \ref{eq:weight_softmax}, and \ref{eq:weightsfull} are still used to calculate the outputted sequence(s) of vectors. 

\subsubsection{Transformers}
Attention mechanisms were usually paired with recurrent neural networks (RNN), such as long short-term memory models \citep{lstm, chang_masterson_2020, xu2015attend}. However, \citet{vaswani.2017} argued that the attention mechanism alone was enough to learn dependencies between words. The architecture built around the attention mechanism is called the transformer. Because the transformer only uses attention, it entirely dispenses with recurrence and more effectively models long-term dependencies between words within the text. In an RNN, words that appear near the beginning of the document may be ``forgotten'' by the end of the document. In (self-)attention, every word is related to every other word, regardless of the distance between words. 

\begin{figure}[!ht]
    \centering
    \includegraphics[scale=0.35]{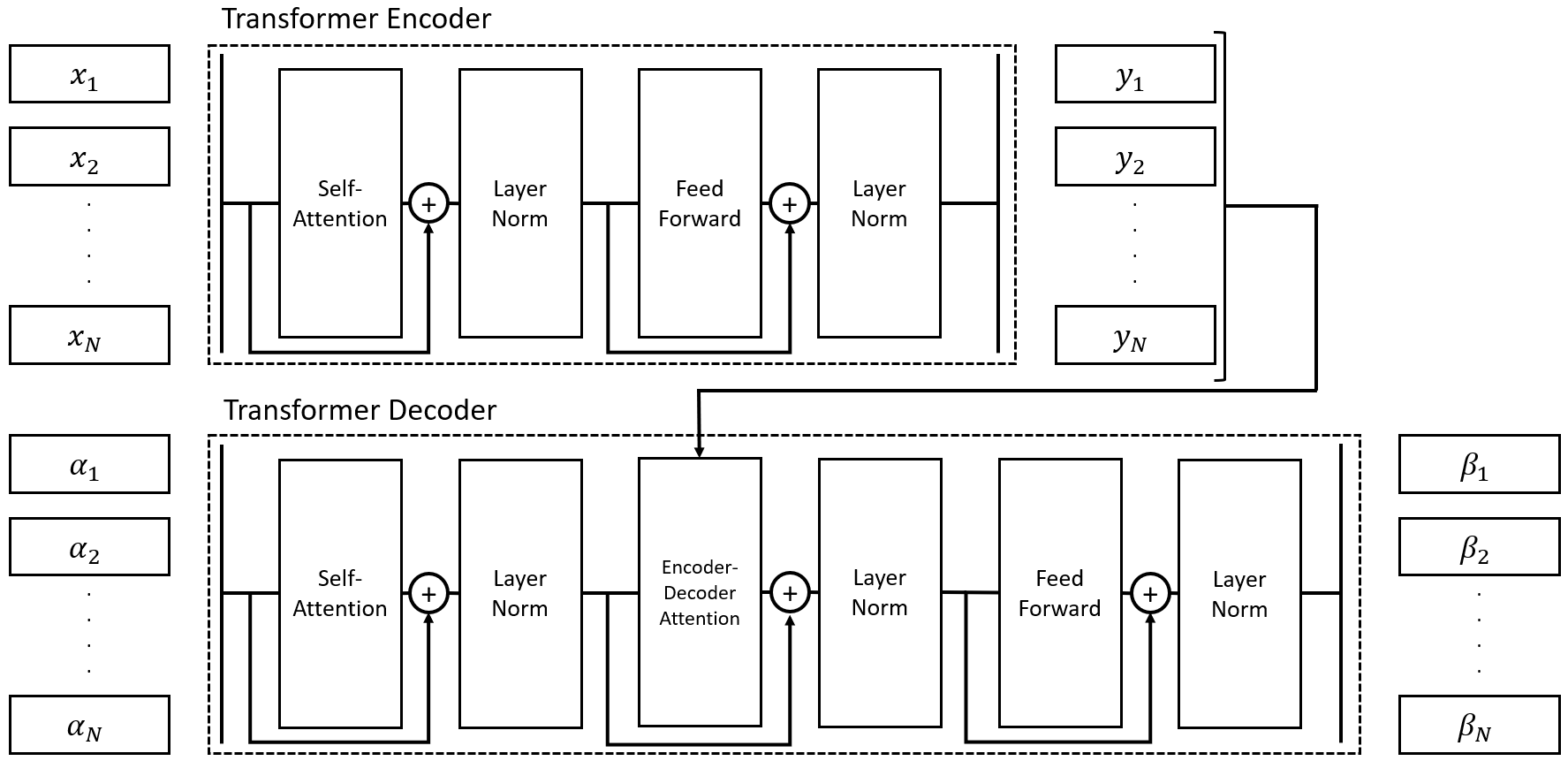}
    \caption[The transformer architecture \citep{vaswani.2017}]{The transformer architecture \citep{vaswani.2017}. The outputs of the transformer encoder are inputted at the encoder-decoder attention layer of the transformer decoder as the keys and values, while the queries comes from the self-attention layer in the transformer decoder. The plus sign with a circle around it indicates a residual connection.}
    \label{fig:transformer}
\end{figure}

The transformer consists of an encoder and a decoder. See Figure \ref{fig:transformer} for an overview of the transformer architecture. Transformers were initially designed for machine translation tasks, where word embeddings $\mathbf{x}$ of one language were inputted into the transformer encoder, and the word embeddings $\mathbf{\alpha}$ of the other language were inputted into the decoder. The output vectors of the transformer encoder $\mathbf{y}$ would be inputted into the encoder-decoder attention layer of the transformer decoder as the keys and values, while the queries came from the self-attention layer of the transformer decoder. 

Many language models, such as BERT, exclusively use the transformer encoder. The transformer encoder is illustrated in the upper half of Figure \ref{fig:transformer}. The transformer encoder consists of a self-attention layer, followed by layer normalization, a feedforward layer where the same feedforward neural network is applied separately to each input, and one last layer normalization; there are residual connections around the self-attention layer and the feedforward layer. 

The transformer decoder, illustrated in the bottom half of Figure \ref{fig:transformer}, exactly resembles the encoder, except for an additional encoder-decoder attention layer and an additional layer normalization and residual connection. The transformer decoder's encoder-decoder attention layer requires the keys and values to come from a separate source, while the output of the self-attention layer in the transformer decoder becomes the queries. 

Notice that the output of the transformer encoder (decoder) can be inputted into another transformer encoder (decoder). This is known as stacking transformer blocks. Most modern architectures stack multiple transformer blocks.

\subsection{Training Early Fusion Models}
\subsubsection{Transfer Learning}
\label{sec:transferlearning}
Deep learning models typically require very large datasets. Thus, one of the principal barriers to using social science data with deep learning methods is labeled data availability. Even if data are readily available, such as social media data, it is still costly to manually annotate the data \citep{webb_williams_casas_wilkerson_2020}. Transfer learning allows researchers to use state-of-the-art deep learning methods with smaller datasets. A formal definition of transfer learning is given in \citet{pan.yang.2010}. 

\begin{figure}[!ht]
    \centering
    \includegraphics[scale=0.5]{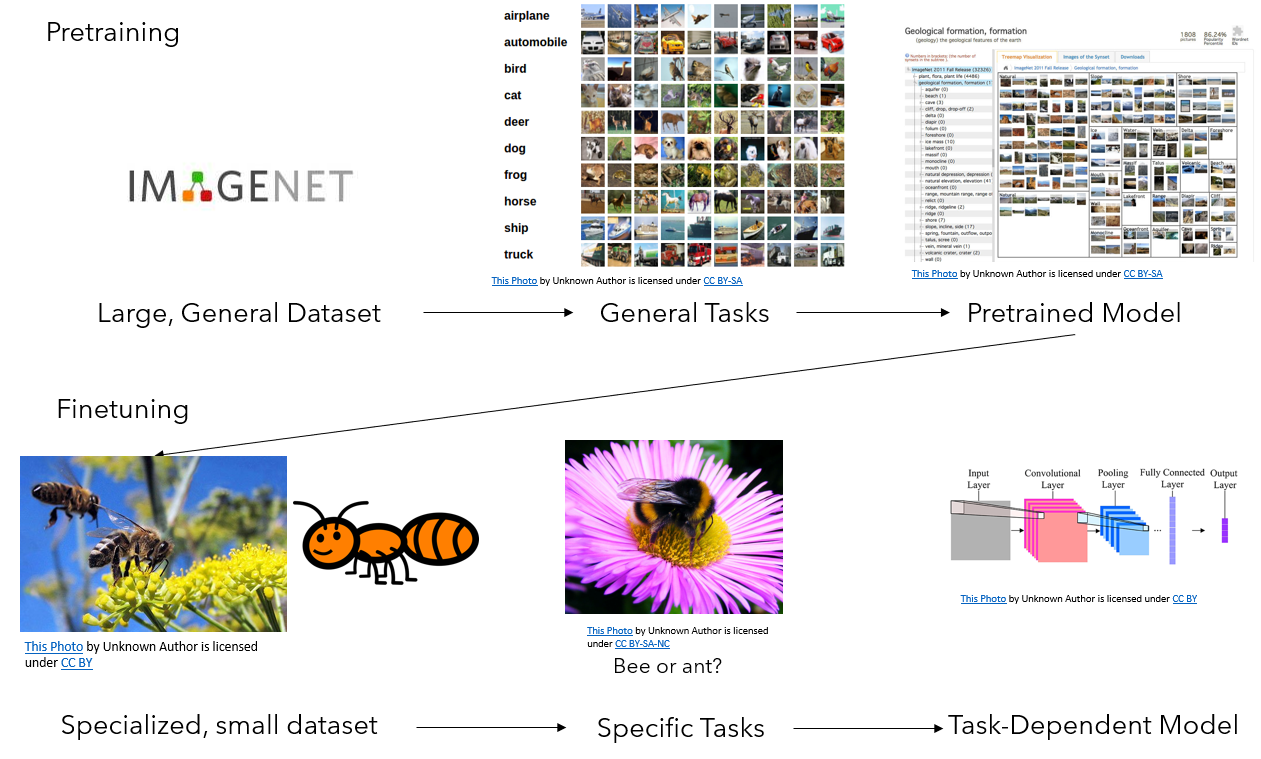}
    \caption[An illustration of the pretraining-finetuning pipeline.]{An illustration of the pretraining-finetuning pipeline classifying images of bees and ants, found in \citet{beesandants}. Even though the training dataset only contains 120 images of bees and ants, the finetuned model is able to correctly classify images of bees and ants with 96\% accuracy.} 
    \label{fig:transfer_learning}
\end{figure}

Informally, transfer learning takes a model trained on a source domain and uses that model to improve the learning of a target predictive function in a separate target domain. Transfer learning typically has two steps: pretraining and finetuning. During \textit{pretraining}, a model is trained to solve general tasks over a large, general dataset. For example, ResNet \citep{he2015deep} is a deep convolutional neural network trained on ImageNet \citep{imagenet}, a dataset of 14 million images with each image belonging to one of 21,841 classes. We train the pretrained model with our data of interest during \textit{finetuning} to learn specific annotation tasks instead of training a model from scratch. Using the pretrained model as the starting point allows us to use deep learning methods on much smaller datasets. This approach is illustrated in Figure \ref{fig:transfer_learning}. Intuitively, transfer learning works because the model learns general concepts during pretraining. For example, ResNet learns shapes, edges, colors, etc., which are visual concepts not exclusive to the images it was originally trained on.

Transfer learning has been most successfully used to pretrain image models, but natural language processing has also recently used the transfer learning paradigm. Pretrained transformers-based language models learn general linguistic concepts such as word order and word similarities; these models are then finetuned for specific downstream tasks \citep{terechshenko_etal_2021}. The most popular of these pretrained language models is bidirectional encoder representations from transformers, or BERT \citep{devlin.chang.lee.toutanova.2018}. BERT follows the transfer learning paradigm described in the previous section and consists of two steps: pretraining and finetuning. First, special tokens are appended to the inputs. A \texttt{[CLS]} token is appended to the beginning of the sentence, while \texttt{[SEP]} is appended at the end of a sentence. The corresponding output vector for \texttt{[CLS]} becomes the sentence or document embedding. \texttt{[SEP]} is used to separate two sentences if two sentences are inputted together into BERT. BERT is pretrained on two tasks during pretraining: the masked language model (MLM) and next sentence prediction (NSP). Details about these pretraining tasks can be found in the Supplemental Information. BERT is pretrained on BooksCorpus (800 million words) and English Wikipedia (2,500 million words). 

\begin{figure}[!ht]
    \centering
    \includegraphics[scale=0.38]{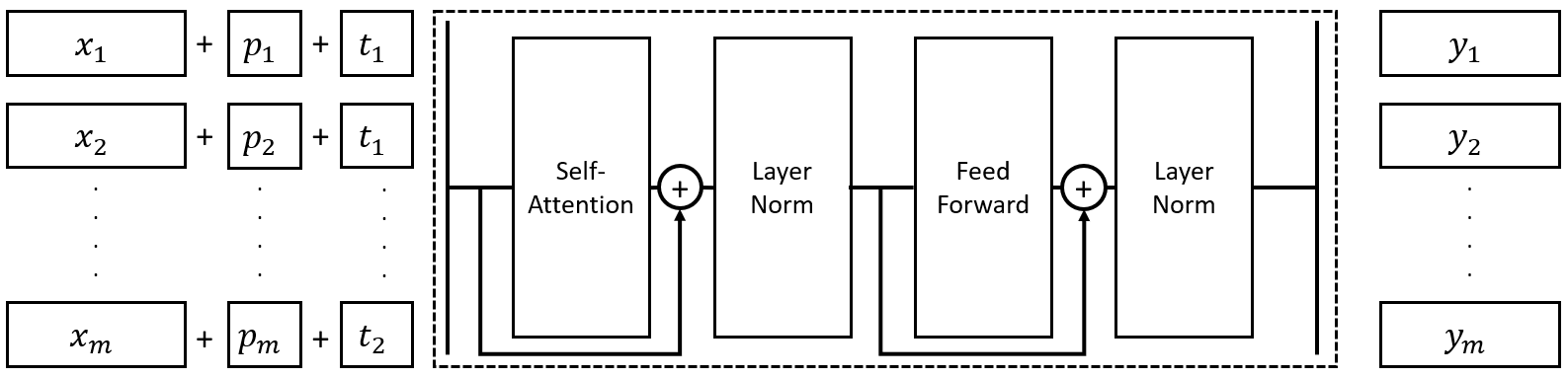}
    \caption[The architecture of BERT \citep{devlin.chang.lee.toutanova.2018}]{The architecture of the first BERT block \citep{devlin.chang.lee.toutanova.2018}. Subsequent BERT blocks do not use position embeddings or token type embeddings. It exactly resembles the transformer encoder, except that token type embeddings and positional embeddings are learned and added to the input. The token type embeddings $\mathbf{t}$ indicate which sentence the embedding comes from; the position embeddings $\mathbf{p}$ take into account word order.} 
    \label{fig:bert}
\end{figure}

Beyond using the \texttt{[SEP]} token between two sentences, BERT also uses token type embeddings in order to distinguish the first and second sentences. Two token type embeddings are learned: $t_1$ and $t_2$. $t_1$ is added to word embeddings that come from the first sentence while $t_2$ is added to word embeddings that come from the second sentence. Positional embeddings, previously discussed in the context of attention, are also used. See Figure \ref{fig:bert} for an overview of a BERT transformer encoder block. 

Pretraining BERT in this fashion, with no labeled dataset but instead using the MLM and NSP pretraining tasks, allows us to train a model that ``understands language'' in a general way. After BERT is pretrained, we can finetune the pretrained weights on a downstream task. Because of the transformer encoder's flexibility to model complexities in language, the parallelizability of transformers, and the ability to capture long-range dependencies between words, almost all state-of-the-art benchmark scores for natural language processing tasks are set by transformer-based pretrained language models.  

\subsubsection{Pretraining Tasks for Early Fusion Multimodal Models}
Because of the success of the transfer learning paradigm in computer vision and natural language processing, it is natural to develop a similar approach for multimodal models. But it is conceptually more difficult to define what ``general'' means in the context of multimodal data. Most state-of-the-art multimodal models are pretrained using image annotation datasets, such as Microsoft COCO \citep{lin2014microsoft} or Conceptual Captions \citep{sharma-etal-2018-conceptual}. Each observation in these datasets contains an image and one or several associated captions in English. General pretraining tasks similar to MLM and NSP used to pretrain BERT are used to pretrain the multimodal models. For example, VisualBERT uses two pretraining tasks: a masked language modeling with the image, where specific tokens of the text must be predicted using the surrounding text and the image, and sentence-image prediction, where the model must predict if a caption actually corresponds with the image or not \citep{li2019visualbert}.

Pretraining multimodal models with image captioning datasets, however, presents a few issues. First, the relationship between an image and its caption is generally not the image-text relationship in real-world observations that use image and text. For example, the text of multimodal social media posts is generally not simply describing what is happening or what objects are in an image. The text and the image modalities extend or modify the overall message the post is attempting to deliver. \citet{singh2020pretraining} find that many of these pretraining tasks do not improve the model's performance. They find that pretraining on data closer to the domain of the downstream task rather than pretraining on image captioning datasets typically yields better performance. However, such experimentation with different pretraining datasets requires computational resources unavailable to most researchers. 

Moreover, even without experimentation, these pretraining tasks are computationally expensive to complete, often requiring the use of several GPUs. Even when pretrained models are available to be downloaded, additional pretraining on the data for the task of interest is required because it allows a model to adapt to a new target domain. State-of-the-art multimodal models are also not developed to accommodate missing modalities. Many models cannot be used with datasets where observations are missing a modality. Others randomly initialize image or text features for observations missing an image or text, respectively.

\section{MARMOT Details}
\begin{figure}[!ht]
    \centering
    \includegraphics[height=3.65in]{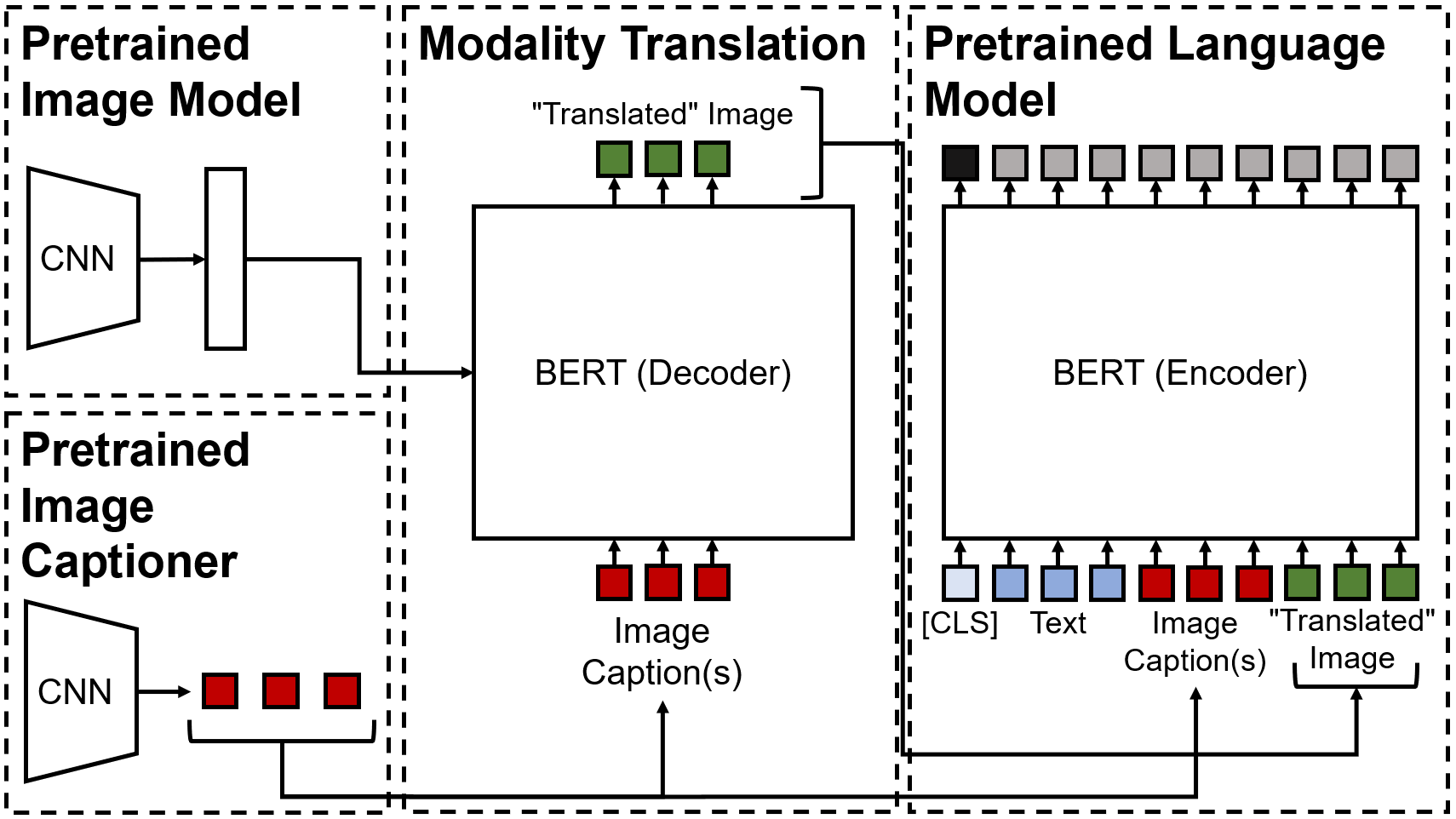}
    \caption[An overview of the MARMOT architecture.]{An overview of the MARMOT architecture. MARMOT uses a pretrained image model to extract image features from the input image. The model flattens these image features and passes them into the encoder-decoder layer of the transformer decoder. Image captions are generated using a pretrained image captioner. During modality translation, the image captions are the input into the BERT decoder. At the encoder-decoder attention layer of the BERT decoder, the image captions attend to the image. The output of the BERT decoder is called the ``translated'' image. This ``translated'' image is then jointly inputted with the text, image captions, and the \texttt{[CLS]} token into the BERT encoder. The joint representation is either the first outputted vector corresponding to the \texttt{[CLS]} token (in black) or the average of the rest of the output vectors (in gray).}
    \label{fig:marmot_diagram}
\end{figure}

Figure \ref{fig:marmot_diagram} shows an overview of the MARMOT architecture. It contains four main components, further described in more detail below: the pretrained image model, the pretrained image captioner, modality translation, and the pretrained language model. The model is simple, and both training and inference can be accomplished using minimal computational resources.\footnote{For example, training and inference can be complete using Google Colab, a free resource that offers the use of one GPU.} The Supplemental Information details what hyperparameters need to be selected, learning rate schedulers, optimizers, and training strategies. 

\subsection{Pretrained Image Model}
The first step involves inputting the image into a pretrained image model, such as ResNet \citep{he2015deep}, Inception v3 \citep{szegedy2015rethinking}, or MobileNet v2 \citep{s2018mobilenetv2}. In our applications, we use ResNet-50 pretrained using the VirTex approach \citep{desai2021virtex}. The VirTex approach pretrains a deep convolutional neural network using image captions instead of labeled images. The Supplemental Information contains more information about the VirTex pretraining approach. We scale down (or up) an arbitrary image $x_{img} \in \mathbb{R}^{3 \times H_0 \times W_0}$ to $\mathbb{R}^{3 \times 224 \times 224}$, and then use ResNet-50 pretrained using VirTex to generate a lower resolution activation map $z \in \mathbb{R}^{2048 \times 7 \times 7}$. Any pretrained image model will work for this step, but exact dimensions may differ.  

\subsection{Pretrained Image Captioner}
At the same time, we generate an image caption for every image in our data. We use self-critical sequence training (SCST) to generate an image caption (or multiple image captions) for each image \citep{rennie2016selfcritical}, although any pretrained image captioner will suffice.\footnote{See the Supplemental Information for more details about SCST.} Generally, multiple captions are used because, during inference, pretrained image captioners may produce different captions that focus on different aspects of the image. 

\subsection{Modality Translation} 
Modality translation captures the spirit of pretraining used in other multimodal models such as VisualBERT \citep{li2019visualbert} without having to further pretrain the model on an image annotations dataset or the data for our task of interest. It also means not having to experiment to find which pretraining dataset works best \citep{singh2020pretraining}. Recall that the goal of pretraining with an image annotations dataset in multimodal models is to adapt the underlying transformer-based pretrained language model (usually BERT) to accept both image and text features and learn patterns between image features and text features. VisualBERT's pretraining tasks reflect these goals: the masked language model with images aims to predict masked out words using the image and the unmasked text, while the sentence-image prediction aims to predict which caption actually belongs to an image \citep{li2019visualbert}. Other models with a pretraining step use similar pretraining tasks.

Instead of pretraining BERT with an image annotations dataset, we provide BERT with explicit image captions generated in the previous step from a pretrained image captioner. A transformer decoder initialized with pretrained BERT weights learns the relationship between the image and the image captions. Inspired by neural translation models (see Figure \ref{fig:transformer}), this step aims to ``translate'' the image features to text features. Simply inputting text and image features directly into a pretrained BERT model without multimodal pretraining is problematic because their input representations have different levels of abstraction \citep{lu2019vilbert}. A learning rate that is too low will better preserve the general language understanding of BERT but learn too little from the image features; a learning rate that is too high will damage the BERT language model's pretrained weights. The image translation step maps the image features to the relevant parts of the text feature subspace to avoid this issue.

To implement image translation, we use a $1 \times 1$ convolution over the activation map $z$ to create a new feature map $z' \in \mathbb{R}^{d \times 7 \times 7}$ (if using ResNet-50 or ResNet-152), where $d=768$ if using $\text{BERT}_{\text{base}}$ or $d=1024$ if using $\text{BERT}_{\text{large}}$. The decoder expects a sequence of vectors as input, so we collapse the spatial dimensions of $z'$ into one dimension, resulting in a $z'' \in \mathbb{R}^{d \times 49}$ feature map. The image feature map $z''$ is inputted at the encoder-decoder layer of the BERT decoder. At the encoder-decoder layer, the image caption attends to the image, decoding the 49 $d$-dimensional vectors in parallel. In essence, this layer re-expresses the word embedding of each word in the image caption as a weighted sum of image features. The output of the BERT decoder is called the ``translated'' image. 

\subsection{Pretrained Language Model} 
The text, image captions, and ``translated'' image feature map are inputted into a transformer initialized with BERT weights. We distinguish text, image captions, and ``translated'' image features using different token type embeddings. Because only two token type embeddings are pretrained by BERT, we initialize the third token type embedding as the average of the two pretrained token type embeddings and add $N(0,0.0001)$ noise to each dimension of the embedding. The model uses 0-indexed position embeddings for each segment, meaning it starts counting from position 0 for each segment. Lastly, we append the \texttt{[CLS]} token to the beginning of the sequence, which acts as an embedding of the observation.

The BERT encoder outputs the joint image-text representation. The representation is either the average across all outputted vectors (the gray outputs in Figure \ref{fig:marmot_diagram}) or the first outputted vector that corresponds to the \texttt{[CLS]} token (the black output in Figure \ref{fig:marmot_diagram}); the choice is a hyperparameter. In our applications, we find the averaging approach typically works slightly better than using the output vector corresponding to the \texttt{[CLS]} token. 

\subsection{Missing Modalities}
To handle missing modalities, we use attention masks. Recall, from Equations \ref{eq:weight_unsoftmaxed}, \ref{eq:weight_softmax}, and \ref{eq:weightsfull}, that the outputted vector $\mathbf{z}_i$ for $i \in \{1,...,N\}$ from attention is calculated as 
\begin{equation*}
    w_{ij}' = \mathbf{q}_i^T \mathbf{k}_j
\end{equation*}
\begin{equation*}
    w_{ij} = \text{softmax}(w_{ij}')
\end{equation*}
\begin{equation*}
    \mathbf{z}_i = \sum_j w_{ij} \mathbf{v}_j
\end{equation*}
We can set the weight $w_{ij}'$ to $-\infty$ for observations of $\mathbf{v}_j$ that need to be excluded from calculating $\mathbf{z}_i$; this effectively sets $w_{ij}$ to 0. This is a mechanism called masking \citep{vaswani.2017}. Masking is typically used for translation tasks when one does not want the attention mechanism to ``peek'' ahead and batching together texts of unequal lengths. 

We take advantage of masking by simply masking out the missing modality in the pretrained language model step. As a concrete example, a dataset may contain observations that all have text, but some are missing an image. We can associate a dummy image and dummy image caption with observations that are missing images. We can mask out the translated dummy image and dummy image caption when they are inputted into the BERT encoder. Backpropagation updates the parameters in the BERT encoder using only the text and does not update the parameters in the BERT decoder in modality translation, as the attention masks block the flow of gradients.

\section{Application 1: Election Incidents Reported on Twitter During the 2016 U.S. General Election}
\subsection{Dataset Background}
\citet{Mebane.Wu.Woods.Klaver.Pineda.Miller2018} collected a dataset of tweets that reported election incidents during the 2016 U.S. general election. An election incident is an individual's personal experience with voting or some other activity in the election. Tweets came between October 1, 2016 and November 8, 2016. The dataset has a binary outcome variable indicating whether the tweet was a reported incident or not. Among the tweets that were classified as an election incident, \citet{Mebane.Wu.Woods.Klaver.Pineda.Miller2018} also include a deeper breakdown about what type of incident it was. The categories are line length/waiting time/polling place overcrowding, polling place event, electoral system, absentee/mail-in/provisional ballot issue, and registration. These categories are then further broken down into subcategories, which are adjectives characterizing the categories. Definitions of the subcategories can be found in the Supplemental Information; exact definitions of the categories can be found in \citet{Mebane.Wu.Woods.Klaver.Pineda.Miller2018}. Human coders of the Twitter data examined all modalities to assign labels. 

In this section, we focus only on the multilabel classification part of the data processing. Of the total 4,018 tweets labeled with categories and subcategories, 1,741 tweets included at least one image. We used 80\% of the dataset for hyperparameter selection and training and set aside the remaining 20\% of the dataset as a test set.

\subsection{Results}
\citet{Mebane.Wu.Woods.Klaver.Pineda.Miller2018} use a text-only ensemble classifier consisting of logistic regression, multinomial naive Bayes, and linear support vector machine. \citet{Mebane.Wu.Woods.Klaver.Pineda.Miller2018} binarize each category and subcategory, meaning that they treat every category and the subcategories under each category as individual binary classification problems. The ensemble classifier only uses text features from the tweets, ignoring all images. \citet{Mebane.Wu.Woods.Klaver.Pineda.Miller2018} also append the date and location of the tweet to the text of the tweet. 

We take the same approach to make results comparable to results found in \citet{Mebane.Wu.Woods.Klaver.Pineda.Miller2018}: we also binarize each category and subcategory and we append the date and location of the tweet to the text of the tweet. MARMOT representations were classified using a two-layer feedforward neural network with a ReLU activation function and we used the cross-entropy loss function. The Supplemental Information contains information about hyperparameters. Results from the ensemble classifier and MARMOT are in Table \ref{tab:electionincidents}; the results are expressed as F1 scores over the positive class. If a tweet had an image, three image captions were generated using self-critical sequence training. 

\begin{table}[!htbp]
\centering
\begin{tabular}{lcc|cc}
\textbf{\textbf{}}                       & \textbf{Ensemble} & \textbf{MARMOT} & \textbf{Support} & \textbf{\# Pictures} \\ \hline
\textbf{Not an Incident}                 & 0.66              & \textbf{0.72}   & 1149             & 330                  \\ \hline
\textbf{Line Length}                     & 0.91              & \textbf{0.92}   & 1045             & 440                  \\
(a) No crowd or no line                  & 0.61              & 0.61            & 85               & 36                   \\
(b) Small crowd or short line            & 0.21              & \textbf{0.30}   & 91               & 31                   \\
(c) Large crowd or long line             & 0.82              & \textbf{0.88}   & 869              & 373                  \\ \hline
\textbf{Polling Place Event}             & 0.78              & \textbf{0.82}   & 1477             & 721                  \\
(a) Did not function as expected         & 0.08              & \textbf{0.49}   & 86               & 19                   \\
(b) Neutral observation                  & 0.47              & \textbf{0.63}   & 481              & 255                  \\
(c) Functioned properly                  & 0.63              & \textbf{0.68}   & 910              & 447                  \\ \hline
\textbf{Electoral System}                & 0.63              & \textbf{0.67}   & 596              & 244                  \\
(a) Did not function properly            & 0.05              & \textbf{0.15}   & 89               & 28                   \\
(b) No comment on function               & 0.65              & \textbf{0.73}   & 473              & 211                  \\ \hline
\textbf{Absentee / Mail-in Voting Issue} & 0.87              & \textbf{0.88}   & 1702             & 748                  \\
(a) Did not function properly            & 0.34              & \textbf{0.53}   & 121              & 28                   \\
(b) Neutral observation                  & 0.60              & \textbf{0.71}   & 613              & 296                  \\
(c) Functioned properly                  & 0.70              & \textbf{0.77}   & 968              & 424                  \\ \hline
\textbf{Registration}                    & 0.85              & \textbf{0.88}   & 475              & 190                  \\
(a) Not able to register                 & 0.17              & \textbf{0.62}   & 59               & 9                   \\
(b) Neutral observation                  & 0.84              & \textbf{0.86}   & 339              & 166                  \\
(c) Able to register                     & 0.50              & \textbf{0.69}   & 77               & 15                  
\end{tabular}
\caption[Binarized classifier performance across the ensemble classifier and MARMOT over the election incidents dataset.]{Binarized classifier performance across the ensemble classifier and MARMOT over the election incidents dataset with 4,018 tweets. Categories are in bold, while subcategories are listed with letters. Results of the ensemble classifier and MARMOT are over a test set that is 20\% of all tweets and the results are F1 scores on the positive class. For a definition of the F1 metric, refer to the Supplemental Information. The total number of observations (across both the training and test sets) that belong to each category and subcategory is noted in the third column. The total number of images (across both the training and test sets) for each category and subcategory is noted in the fourth column. The results of the ensemble classifier come directly from \citet{Mebane.Wu.Woods.Klaver.Pineda.Miller2018}. Numbers are rounded to two decimal places because the results of the ensemble classifier were originally reported to only two digit places.} 
\label{tab:electionincidents}
\end{table}

MARMOT outperforms the ensemble classifier on all categories and subcategories except for ``Line Length: No crowd or no line.'' In that specific subcategory, MARMOT matches the performance of the text-only ensemble classifier. There are improvements in the F1 scores where we would intuitively expect images to play a role, such as in the short and long line subcategories under the ``Line Length'' category. There are also improvements in some subcategories that contain few images. For example, the text-only ensemble classifier struggled with the subcategory ``Polling Place Event: Did not function as expected.'' MARMOT performed significantly better, despite the subcategory containing very few images: only 19 of the 86 tweets in this subcategory had an image. In other words, it is not immediately apparent that all improvements are the direct result of including images. MARMOT uses BERT, which consistently outperforms other text classifiers as well \citep{devlin.chang.lee.toutanova.2018}. We turn to look at model variants to examine this possibility. 

\subsection{Model Variants}
We look at two variants of the model: the first uses only the text from each observation with the standard BERT encoder, and the second uses the text and image captions with the BERT encoder but does not use the translated image. The results of the two model variants, along with the full MARMOT model, are in Table \ref{tab:ablations_electionincidents}, which details the F1 score over the positive class for each model variant.

\begin{table}[!htbp]
\centering
\begin{tabular}{lccc}
\textbf{}                                & \Centerstack{ \textbf{BERT,} \\ \textbf{Text Only}} & \Centerstack{ \textbf{BERT,} \\ \textbf{Text and} \\ \textbf{Image Captions}} & \textbf{MARMOT} \\ \hline
\textbf{Not an Incident}                 & 0.65                         & 0.71                                       & 0.72            \\ \hline
\textbf{Line Length}                     & 0.92                         & 0.92                                       & 0.92            \\
(a) No crowd or no line                  & 0.44                         & 0.44                                       & 0.61            \\
(b) Small crowd or short line            & 0.21                         & 0.19                                       & 0.30            \\
(c) Large crowd or long line             & 0.86                         & 0.87                                       & 0.88            \\ \hline
\textbf{Polling Place Event}             & 0.81                         & 0.81                                       & 0.82             \\
(a) Did not function as expected         & 0.36                         & 0.41                                       & 0.49            \\
(b) Neutral observation                  & 0.55                         & 0.58                                       & 0.63            \\
(c) Functioned properly                  & 0.65                         & 0.65                                       & 0.68            \\ \hline
\textbf{Electoral System}                & 0.66                         & 0.65                                       & 0.67            \\
(a) Did not function properly            & 0.11                         & 0.11                                       & 0.15            \\
(b) No comment on function               & 0.72                         & 0.72                                       & 0.73            \\ \hline
\textbf{Absentee / Mail-in Voting Issue} & 0.87                         & 0.87                                       & 0.88            \\
(a) Did not function properly            & 0.42                         & 0.49                                       & 0.53            \\
(b) Neutral observation                  & 0.67                         & 0.67                                       & 0.71            \\
(c) Functioned properly                  & 0.77                         & 0.77                                       & 0.77            \\ \hline
\textbf{Registration}                    & 0.89                         & 0.87                                       & 0.88            \\
(a) Not able to register                 & 0.50                         & 0.52                                       & 0.62            \\
(b) Neutral observation                  & 0.88                         & 0.85                                       & 0.86            \\
(c) Able to register                     & 0.65                         & 0.63                                       & 0.69         
\end{tabular}
\caption[Results of the MARMOT model variants over the election incidents dataset.]{Results of MARMOT model variants over the election incidents dataset with 4,018 tweets. The first column reports the F1 scores over each binarized category or subcategory from a model using a pretrained BERT encoder with only text input. The second column reports the results from a model using a pretrained BERT encoder with text and image captions as inputs, but does not use the translated image. The third column reports the results from the full MARMOT model.}
\label{tab:ablations_electionincidents}
\end{table}

We can attribute many of the improvements over the text-only ensemble classifier baseline to BERT. For example, most of the improvements in the category ``Line Length: Large crowd or long line'' came from BERT. The inclusion of images did offer an improvement in the F1 score in several subcategories---namely, ``Not an Incident,'' ``Line Length: No crowd or no line,'' ``Line Length: Small crowd or short line,'' ``Polling Place Event: Did not function as expected,'' ``Polling Place Event: Functioned properly,'' ``Electoral System: Did not function properly,'' ``Absentee / Mail-in Voting Issue: Did not function properly,'' ``Absentee / Mail-in Voting Issue: Neutral observation,'' and ``Registration: Able to register.'' Therefore, not all performance gains over the text-only ensemble classifier baseline resulted from using a more powerful text representation architecture. 

The \textit{lack} of an image may help MARMOT to predict many more tweets correctly. Recall that MARMOT deals with a missing image in an observation by masking out a dummy image. The learned representation is different for observations that only have text versus observations with both image and text. The differences in these representations can be a pattern differentiating a positive classification from a negative classification. MARMOT may be helpful in both situations where images play a role in an observation's classification and situations where the lack of an image may further clarify an observation's classification.  

\subsection{Examples of Successful Classifications}
We take a qualitative look at some of the classifier results using the MARMOT representations to understand better the strengths of the MARMOT multimodal representations. 

\subsubsection{Line Length: Large Crowd or Long Line}
\begin{figure}[!htbp]
\minipage{0.5\textwidth}
  \centering
  \includegraphics[height=4in,clip]{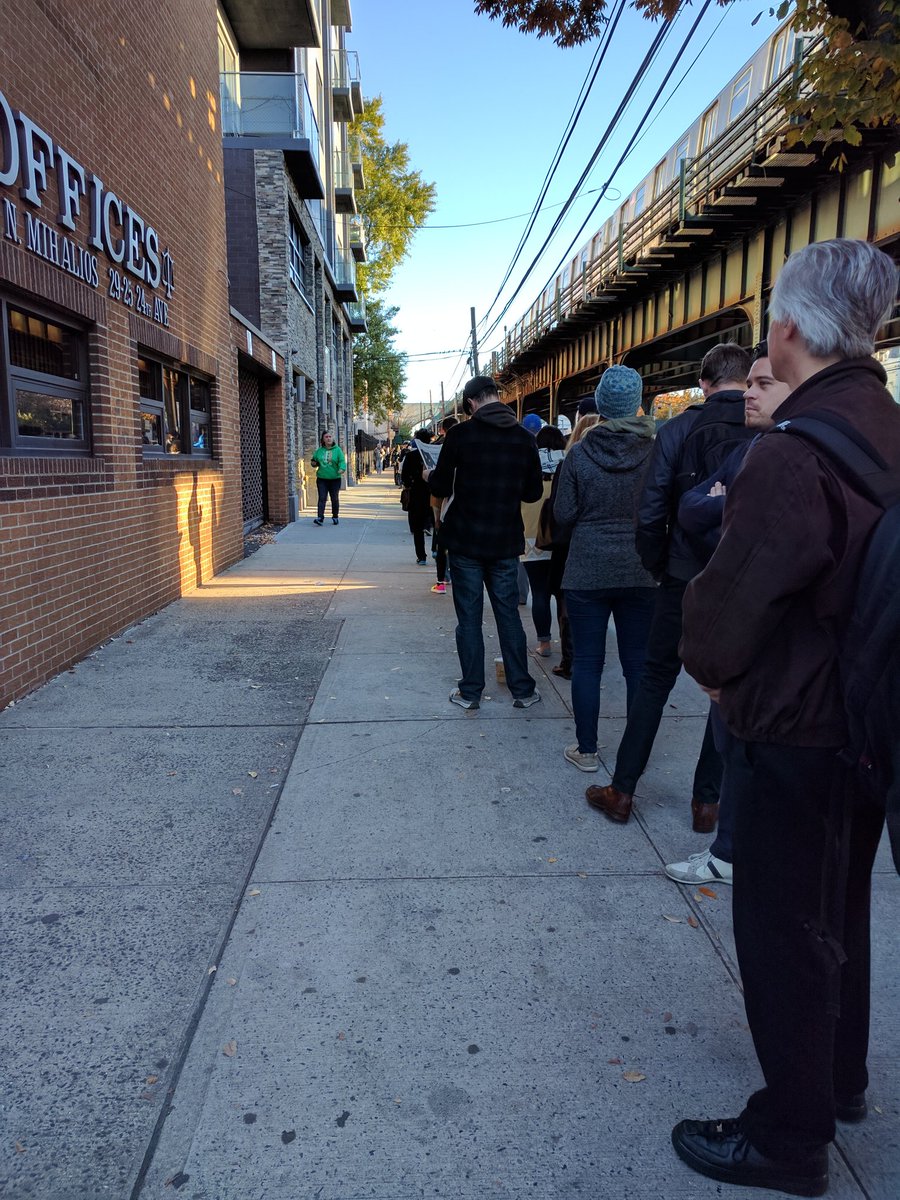}
  \caption*{``Line to vote in astoria queens.''}
\endminipage\hfill
\minipage{0.5\textwidth}
  \centering
  \includegraphics[height=3.8in,clip]{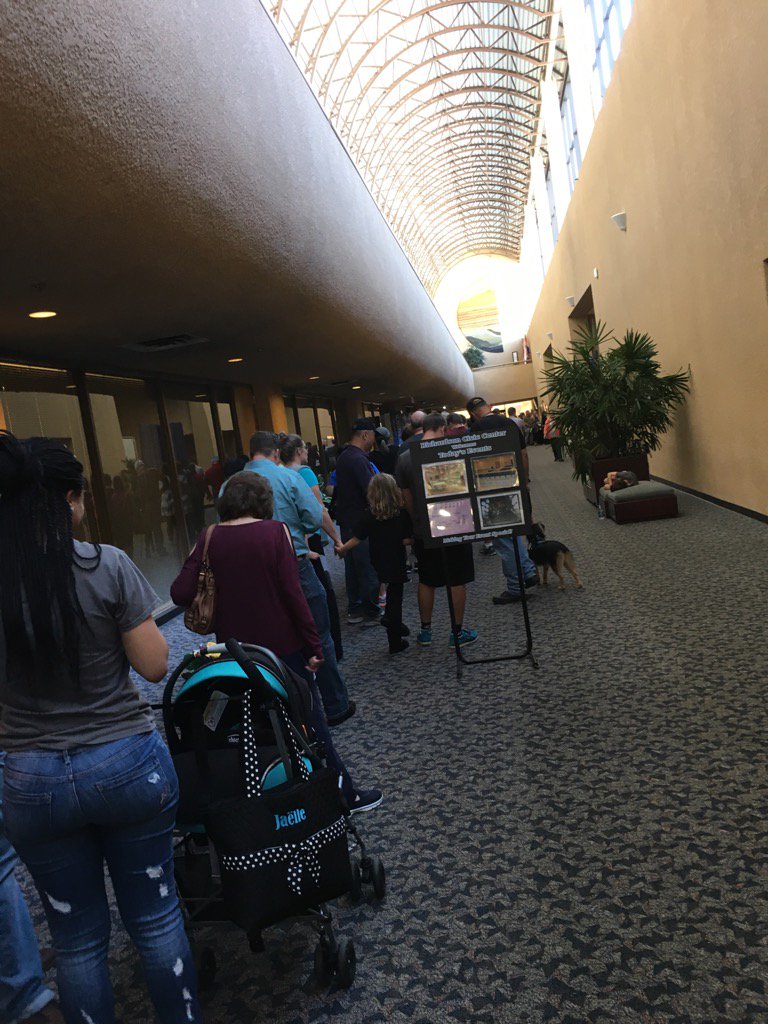}
  \caption*{``The line for first day of early voting, Richardson TX''}
\endminipage
\caption{Two example tweets of long lines at polling stations.} 
\label{fig:longlines}
\end{figure}

MARMOT outperformed the text-only ensemble classifier in the large crowd or long line subcategory. Figure \ref{fig:longlines} show examples of tweets that were correctly classified as long lines at polling places by MARMOT. 

\begin{figure}[!htbp]
    \centering
    \includegraphics[scale=0.38]{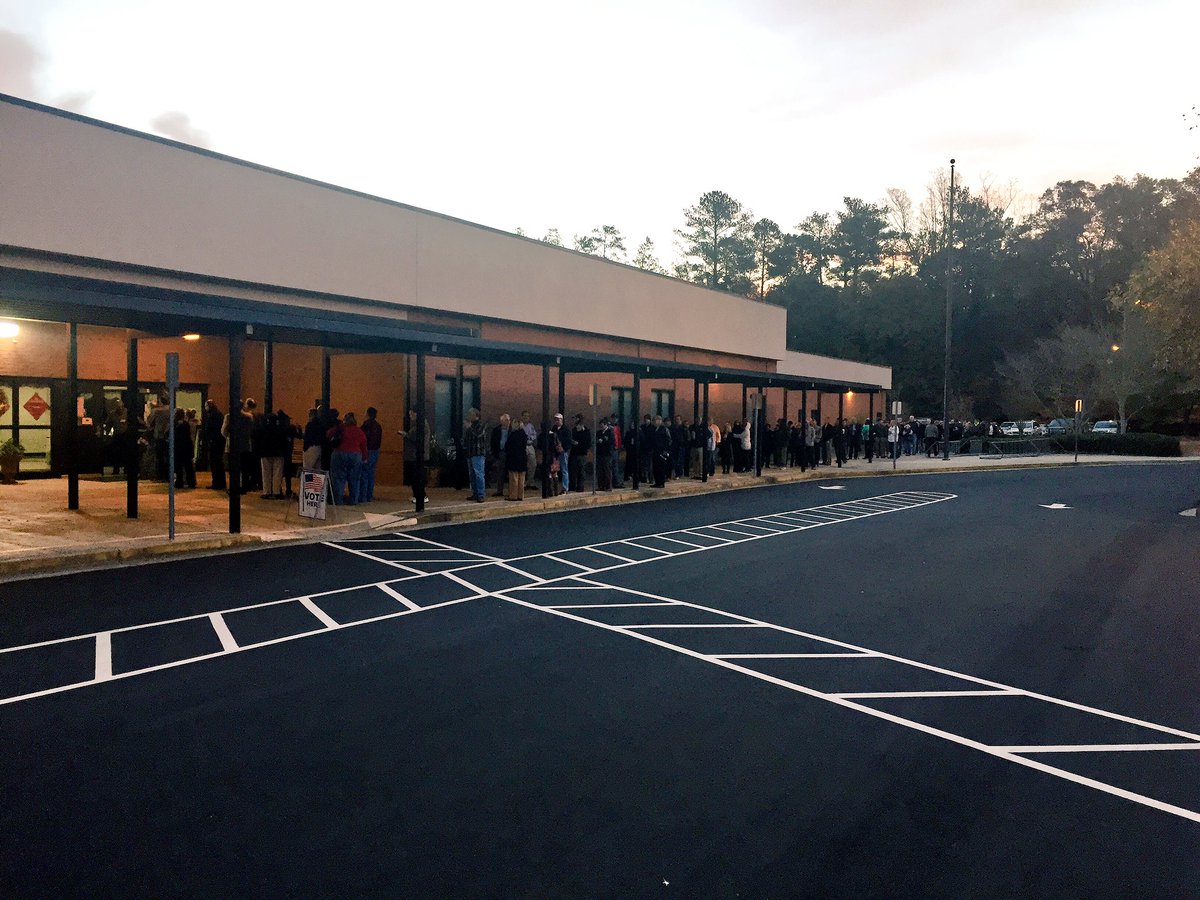}
    \caption{``Polls are open in GA. Proud to cast my ballot. Longest line ever seen in Peachtree Corners!''}
    \label{fig:line_peach}
\end{figure}

The text-only BERT classifier, however, also correctly classified most of these types of examples. MARMOT only did marginally better in this subcategory in terms of the F1 score. Plausibly, the text-only models can learn to classify any tweet mentioning a line at a polling station as a report of a long line at a polling station. However, we can see that MARMOT is slightly more nuanced than that. Figure \ref{fig:line_peach} is an example of a tweet that reports a long line at a polling place. It notes that the polls are open in Georgia, but it did not refer to Peachtree Corners as a polling place. The image depicts a scene where there is a long line at a polling station, indicated by the ``Vote Here'' sign. The text-only BERT classifier misclassified this tweet, while MARMOT correctly classified the tweet as depicting a long line at a polling place. 

\subsubsection{Polling Place Events: Functioned Properly}
\begin{figure}[!htbp]
\minipage{0.45\textwidth}
  \centering
  \includegraphics[height=2.9in]{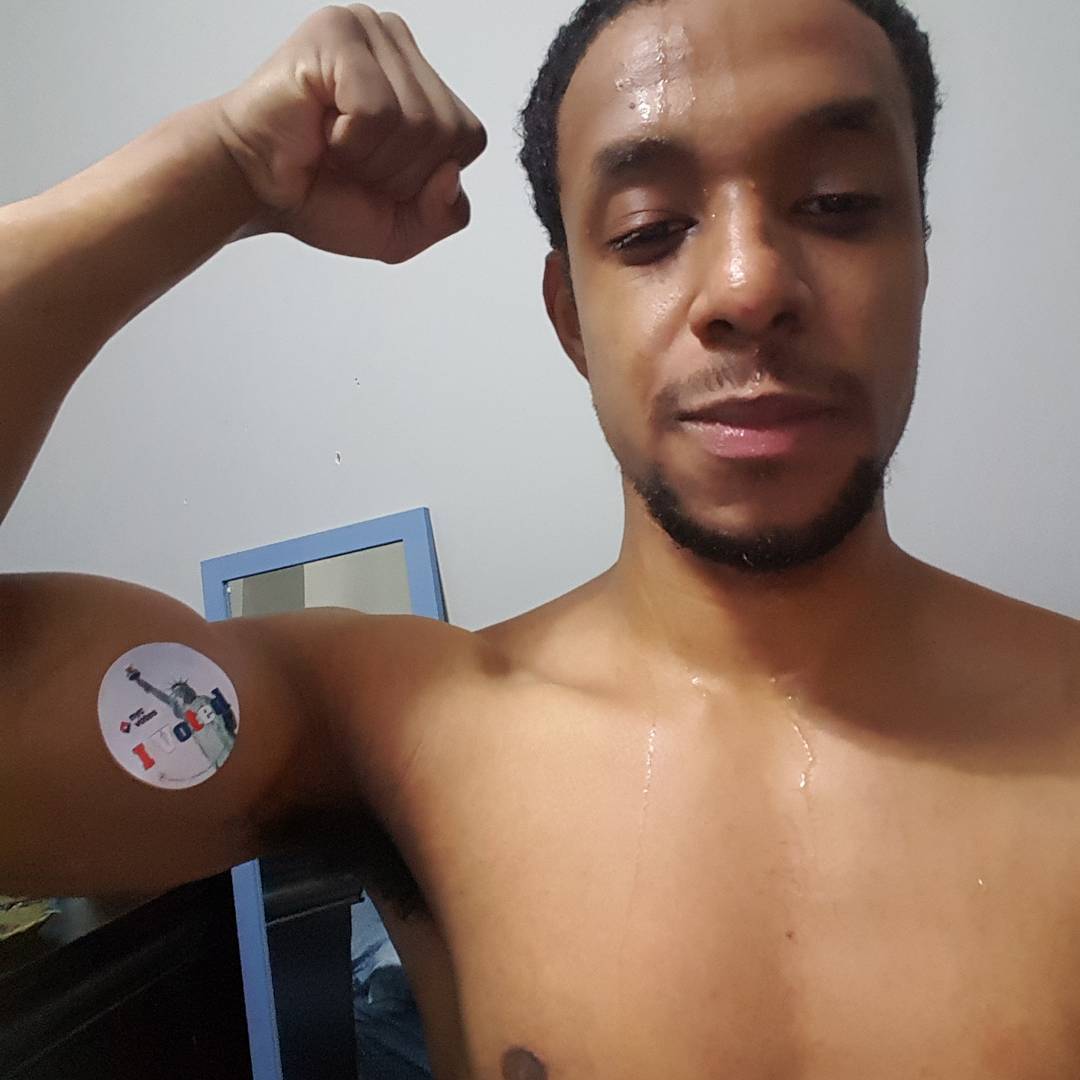}
  \caption*{``\#Voting makes you \#Stronger. \#Vote The work doesn't stop after today.''}
\endminipage\hfill
\minipage{0.45\textwidth}
  \centering
  \includegraphics[height=2.7in]{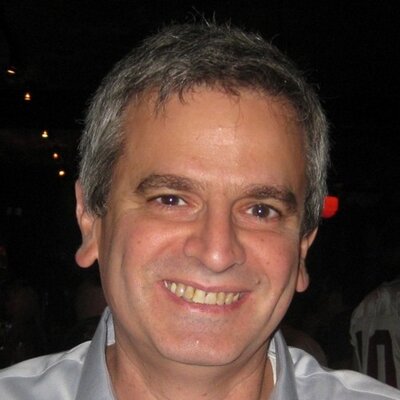}
  \caption*{``I felt the same way as I arrived to the polling station. As an immigrant, Im honored to be able to vote.''}
\endminipage
\caption{Two example tweets of polling places functioning correctly.}
\label{fig:ppe}
\end{figure}

Images play a role in classifying tweets of polling places functioning correctly, particularly tweets involving people smiling after voting. Figure \ref{fig:ppe} contains two example tweets of polling places functioning correctly. In the first example, the person advocates for voting but does not directly indicate with the text that he successfully voted. The image depicts a person with an ``I Voted'' sticker on his arm, indicating that he was successful in voting and thus that a polling place functioned correctly. The text-only classifier failed to classify this observation as polling place functioning correctly, while MARMOT correctly classified this tweet as a tweet indicating a polling place functioning correctly. In the second example, the individual reports that he was honored to vote and included a picture of himself smiling. The text indicates that he voted at a polling station but does not describe his experience; the picture, on the other hand, indicates that he was happy to vote, suggesting that the polling place functioned correctly. MARMOT correctly classifies this tweet as a tweet indicating a polling place functioning correctly; the text-only classifier, however, misclassified this tweet.  

\section{Application 2: Hateful Memes}
There is increasing interest in social sciences and computer science around detecting and analyzing hate speech on social media. \citet{trumphateontwitter} looked at 750 million tweets and did not find an increase in hate speech or white nationalistic language on Twitter. \citet{siegel_badaan_2020} examine what counter-speech initiatives are most effective at reducing sectarian hate speech online. \citet{macavaneyetal} and \citet{davidson2017automated} examine the machine learning approaches to detecting hate speech on social media. 

The Hateful Memes dataset aims to help develop models that more effectively detect multimodal hateful content. Besides being a well-curated dataset for building models that detect multimodal hate speech, the Hateful Memes dataset is also useful for comparing how MARMOT performs against the state-of-the-art multimodal models because every observation has both text and image. We find that MARMOT shows significant improvements over the results of benchmark state-of-the-art multimodal models on this classification problem, suggesting that MARMOT does not lack performance over other multimodal classifiers even though it does not require pretraining on an image annotations dataset and it can calculate representations for observations missing modalities.

\subsection{Dataset Background}
Facebook Research recently released the Hateful Memes dataset to develop and test multimodal models \citep{kiela2020hateful}. This dataset was also used in the Hateful Memes challenge.\footnote{For more information about this challenge, see \url{https://ai.facebook.com/blog/hateful-memes-challenge-and-data-set/}.} \citet{kiela2020hateful} define hate as follows: ``A direct or indirect attack on people based on characteristics, including ethnicity, race, nationality, immigration status, religion, caste, sex, gender identity, sexual orientation, and disability or disease. We define attack as violent or dehumanizing (comparing people to non-human things, e.g. animals) speech, statements of inferiority, and calls for exclusion or segregation. Mocking hate crime is also considered hate speech.'' 

The problem requires a multimodal model to solve because the image and text may, on their own, not be hateful, but the combination of the two may be hateful. To give an example of an unkind (but not hateful) meme that captures this idea, consider an image of a skunk paired with the text, ``Love the way you smell today.'' Neither the text nor the image is mean on its own. However, the combination of the two modalities makes it an unkind meme. 

The dataset contains 10,000 memes. The validation set is 5\% of the data, the test set is 10\% of the data, and the rest of the data is set aside as training data. Most memes were actually found on Facebook, but they also created a set of benign confounders, which are artificially-created memes based on an actual hateful meme that has been made non-hateful by replacing the image or replacing the text. In total, there are five types of memes in the dataset: multimodal hate, where the text and image on their own are not hateful but are hateful when paired together; unimodal hate, where the text or image (or both) are already hateful on their own; benign text confounders; benign image confounders; and random non-hateful examples. Multimodal hate makes up 40\% of the data, unimodal hate makes up 10\% of the data, benign text confounders make up 20\% of the data, benign image confounders make up 20\% of the data, and 10\% of the data are random non-hateful. The dataset, however, does \textit{not} specify which observations belong to which categories, meaning we could not use the type of meme as part of the classification process.\footnote{The noted existence of these meme category types was a significant issue during the Hateful Memes challenge. Some contestants designed architectures to predict the category of each meme in the dataset. These category-specific architectures led to extraordinary performances that far exceeded human coder baselines. As these category types would not exist in any real-life setting involving memes, our approach does not attempt to infer the category types of each meme as part of the prediction pipeline.} The outcome of interest is whether a meme is hateful or not. The Hateful Memes dataset has 5,000 hateful memes and 5,000 non-hateful memes. 

\subsection{Results}
For MARMOT, we used a two-layer feedforward neural network with a ReLU activation function as the classifier and we used the cross-entropy loss function. We generated three captions for each image using self-critical sequence training. For information about hyperparameter selection, see the Supplemental Information. The accuracy learning curve of MARMOT over the validation set, used to assess potential overfitting, can be found in the Supplemental Information. \citet{kiela2020hateful} provide results over the test set for many multimodal models, including ViLBERT, VisualBERT, and MMBT, which are state-of-the-art multimodal classifiers. Results over the test set can be found in Table \ref{tab:hatefulmemes}. Results over the validation set can be found in the Supplemental Information. 

\begin{table}[!htbp]
\centering
\begin{tabular}{l|cc}
    \textbf{Model}  & \textbf{Accuracy} & \textbf{AUC}    \\ \hline
    Image - Grid    & 0.5200            & 0.5263          \\
    Image - Region  & 0.5213            & 0.5592          \\
    Text BERT       & 0.5920            & 0.6508          \\
    Late Fusion     & 0.5966            & 0.6475          \\
    Concat BERT     & 0.5913            & 0.6579          \\
    MMBT - Grid     & 0.6006            & 0.6792          \\
    MMBT - Region   & 0.6023            & 0.7073          \\
    ViLBERT         & 0.6230            & 0.7045          \\
    VisualBERT      & 0.6320            & 0.7133          \\
    ViLBERT CC      & 0.6110            & 0.7003          \\
    VisualBERT COCO & 0.6473            & 0.7141          \\
    MARMOT          & 0.6760   & \textbf{0.7530} \\
    MARMOT - Deep Ensemble & \textbf{0.6920} & 0.7493 
    \end{tabular}
    \caption[Accuracy and area under the receiver operating characteristic curve (AUC) performance metrics across the 11 baseline models, MARMOT, and MARMOT in a deep ensemble over the test set of the Hateful Memes dataset.]{Accuracy and area under the receiver operating characteristic curve (AUC) performance metrics across the 11 baseline models, MARMOT, and MARMOT in a deep ensemble over the test set of the Hateful Memes dataset. For definitions of these metrics, refer to the Supplemental Information. ``Grid'' means that image features derived from ResNet-152 were used as input features. ``Region'' means that segmented image features derived using Faster R-CNN \citep{ren2015faster}, rather than the entire image, were used as input features. Concat BERT means that BERT embedding features were concatenated with image features. Results are over the test set, which is 1,000 memes. VisualBERT COCO means that VisualBERT was pretrained over the MS COCO dataset \citep{lin2014microsoft}. ViLBERT CC means ViLBERT was pretrained over the Conceptual Captions dataset \citep{sharma-etal-2018-conceptual}. ``Deep Ensemble'' means that the final prediction for each observation was the majority predicted class across 11 iterations of MARMOT.} 
    \label{tab:hatefulmemes}
\end{table}

MARMOT outperforms the benchmark state-of-the-art multimodal classifiers on both accuracy and area under the receiver operating characteristic curve (AUC).\footnote{Intuitively, AUC is the probability that a classifier will rank a randomly chosen positive example over a randomly chosen negative example. For a more detailed set of definitions of the performance metrics, refer to the section on evaluation metrics in the Supplemental Information.} MARMOT also outperforms the pretrained variants of ViLBERT and VisualBERT. MARMOT does not trade off performance on classification problems to accommodate potentially missing modalities---it can perform just as well, if not better, than the baseline state-of-the-art multimodal models while still retaining this critical property that makes it useful for social science research. MARMOT's accuracy improves further when MARMOT is used in a deep ensemble, which uses the majority predicted class across 11 separate iterations of MARMOT \citep{lakshminarayanan2017simple}. In the Hateful Memes challenge, MARMOT finished in the top 1\% of all contestants.\footnote{Placement in the challenge was based on AUC.} 

\subsection{Model Variants}
We analyze two variants of the model: the first uses only the text from each observation with BERT and the second uses the text and image captions with BERT but not the translated image. The results of the two model variants are in Table \ref{tab:ablationstudies_hm}. 

\begin{table}[!htbp]
\centering
\begin{tabular}{l|ll}
\textbf{Model Variant}                  & \textbf{Accuracy} & \textbf{AUC} \\ \hline
BERT with Text Only               & 0.5920            & 0.6508       \\
BERT with Text and Image Captions & 0.6510            & 0.7469       \\
MARMOT                            & 0.6760            & 0.7530       \\
MARMOT - Deep Ensemble            & 0.6920            & 0.7493
\end{tabular}
\caption[Results of the MARMOT model variants over the test set of the Hateful Memes dataset.]{Results of the MARMOT model variants over the test set of the Hateful Memes dataset. The first model variant uses only the text; the result is taken directly from \citet{kiela2020hateful}. The second model variant uses the text and image captions generated during modality translation, but the translated image is not used. The third model variant is the full MARMOT model. The fourth model variant is MARMOT used in a deep ensemble.}
\label{tab:ablationstudies_hm}
\end{table}

The addition of the generated image captions accounts for most of the improvement over using the text alone with a BERT model: there is a 9 point improvement in AUC and a 6 point improvement in accuracy with the image captions. As the captions are text, BERT can easily learn the relationships between the text and the image captions. There is a further improvement when using the full MARMOT model, especially in terms of accuracy. Image translation can capture additional information about the image that is useful for the detection of hateful memes. 

\subsection{Examples of Successful Classifications}
To more intuitively understand how MARMOT is classifying the memes as hateful or not, we qualitatively look at a few classification examples where MARMOT was successful. Please note that some of these examples may contain sensitive or offensive content. 

\begin{figure}[!htbp]
    \centering
    \includegraphics[scale=0.6]{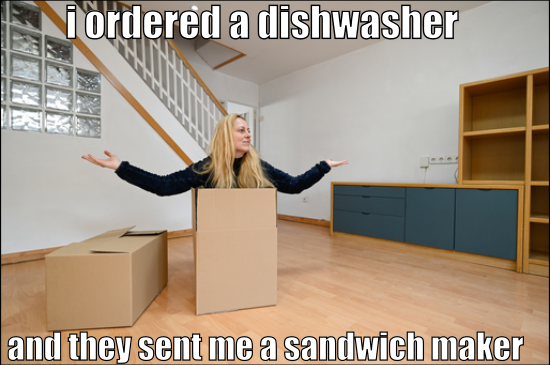}
    \caption[An example of a multimodal hateful meme.]{An example of a multimodal hateful meme. This meme was correctly classified by MARMOT as hateful. \copyright Getty Images.}
    \label{fig:multimodal_hateful}
\end{figure}

The most difficult memes to classify as hateful are the multimodal hate memes. These are the memes where the text alone is not hateful, and the image alone is not hateful, but when put together the meme becomes hateful. Figure \ref{fig:multimodal_hateful} shows an example of a multimodal hateful meme which MARMOT correctly classified as hateful. 

\begin{figure}[!htbp]
\minipage{0.333\textwidth}
  \centering
  \includegraphics[width=2.1in]{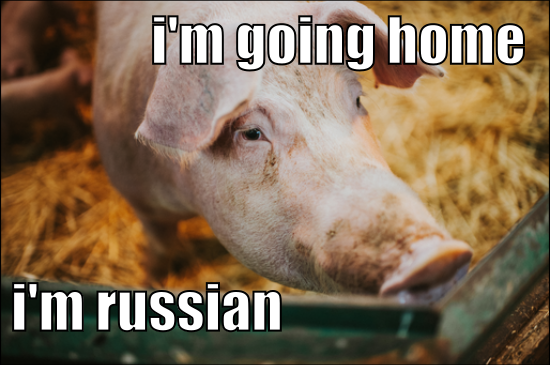}
\endminipage\hfill
\minipage{0.333\textwidth}
  \centering
  \includegraphics[width=2.1in]{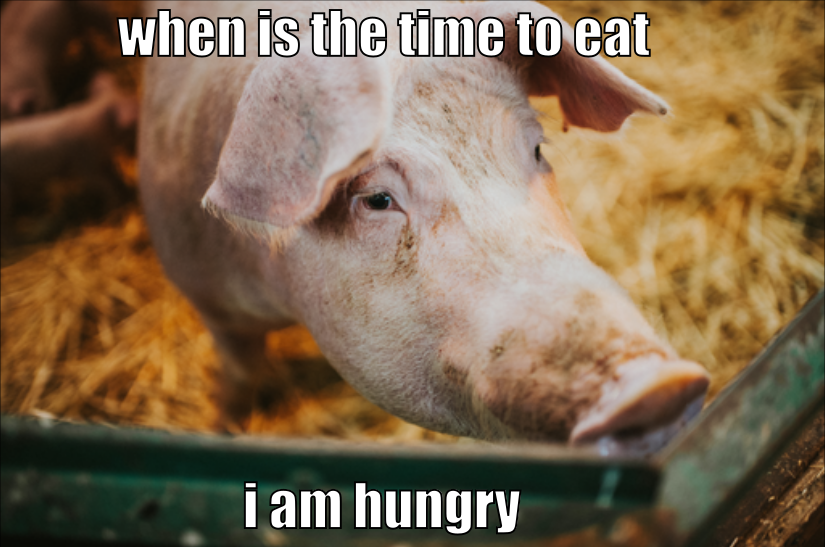}
\endminipage \hfill
\minipage{0.333\textwidth}
  \centering 
  \includegraphics[width=2.1in]{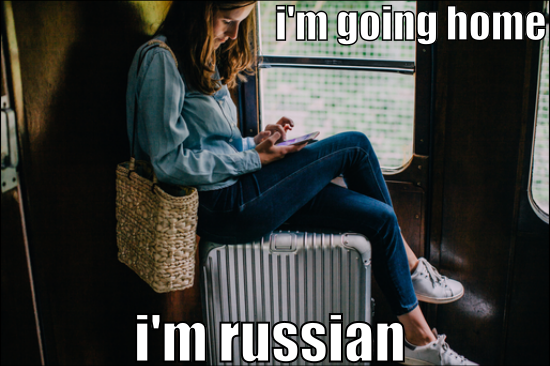}
\endminipage 
\caption[An example of a multimodal hateful meme with a benign text confounder counterpart and benign image confounder counterpart.]{An example of a multimodal hateful meme with a benign text confounder counterpart and benign image confounder counterpart. MARMOT correctly classifies the meme on the left as hateful and the meme in the center and on the right as non-hateful. \copyright Getty Images.}
\label{fig:triplet1}
\end{figure}

The model may learn that the use of the words ``dishwasher'' or ``sandwich maker'' in a memes setting usually implies that the meme is misogynistic. After all, across the training and validation datasets, most of the memes containing the phrase ``dishwasher'' or ``sandwich maker'' are hateful. To further demonstrate evidence that the model is learning relationships between the image and the text, we first look at an example, shown in Figure \ref{fig:triplet1}, consisting of three memes. The memes on the left and in the center have the same image but different text (benign text confounder). The memes on the left and the right have different images but the same text (benign image confounder). MARMOT correctly classifies the first meme on the left as hateful and correctly classifies the meme in the center and on the right as non-hateful.\footnote{Specifically, the probability of each meme being hateful, using MARMOT, is 0.929, 0.0008, and 0.013, respectively.}

\begin{figure}[!htbp]
\minipage{0.333\textwidth}
  \centering
  \includegraphics[width=2.15in]{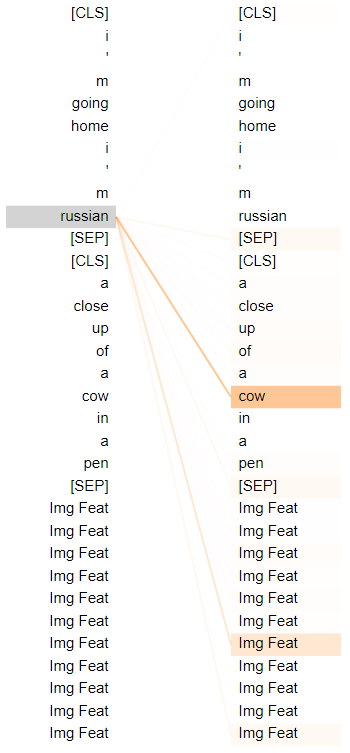}
\endminipage\hfill
\minipage{0.333\textwidth}
  \centering
  \includegraphics[width=2.15in]{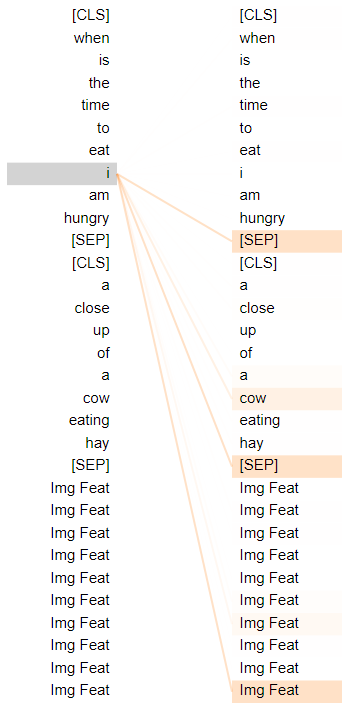}
\endminipage\hfill
\minipage{0.333\textwidth}
  \centering
  \includegraphics[width=2.15in]{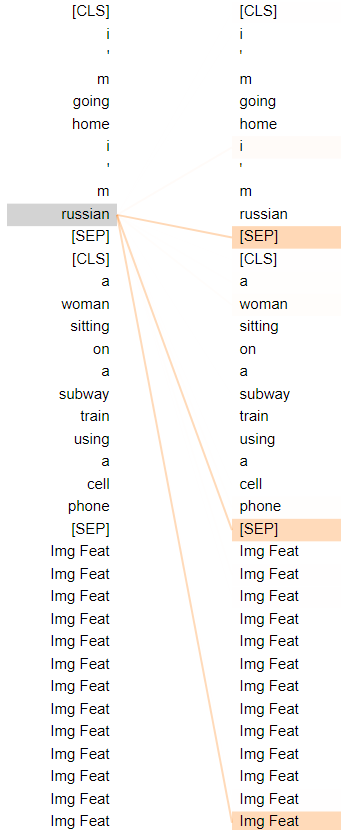}
\endminipage
\caption[A visualization of the attention weights from the 12th attention head of the 10th layer of the BERT encoder from the last step of MARMOT for the three memes from Figure \ref{fig:triplet1}.]{A visualization of the attention weights from the 12th attention head of the 10th layer of the BERT encoder from the last step of MARMOT for the three memes from Figure \ref{fig:triplet1}. The visualizations on the left, center, and right correspond with the left, center, and right memes, respectively, from Figure \ref{fig:triplet1}. The line weight indicates the attention weight. Visualizations were created using the package described in \citet{vig2019transformervis}. The token ``Img Feat'' indicates an inputted image feature, which does not translate to a word token.} 
\label{fig:triplet_attentionweights}
\end{figure}

To take this analysis one step further, we create visualizations of the attention weights of the BERT encoder used in the last stage of MARMOT.\footnote{Because of computational constraints, these visualizations could not be created for tweets of election incidents during the 2016 U.S. general election.} To be clear, these visualizations do not causally show why MARMOT correctly classified a specific observation; in other words, we are not trying to learn about intent or meaning behind how the MARMOT representations are constructed. Instead, these visualizations show that MARMOT can learn important associations between the image and text. Figure \ref{fig:triplet_attentionweights} shows the attention weights of the 12th attention head of the 10th transformer layer in the BERT encoder. The line weight reflects the attention weight between the attending tokens and the attended tokens. In the hateful example (the meme on the left in Figure \ref{fig:triplet1}), the token ``russian'' from the text of the meme shares a high attention weight with the token ``cow'' from the image caption.\footnote{Self-critical sequence training misidentifies the animal in the picture as a cow.} In the non-hateful benign text confounder example (the meme in the center in Figure \ref{fig:triplet1}), the token ``i'' from the text of the meme shares a lower attention weight with the token ``cow'' from the image caption compared to the hateful example. In the non-hateful benign image confounder example (the meme on the right in Figure \ref{fig:triplet1}), there is essentially no attention weight learned between the token ``russian'' and the image caption. The attention weights suggest that MARMOT learns different relationships among the images and text between the hateful and non-hateful examples, even when the text or image is the same. 

\begin{figure}[!htbp]
\minipage{0.5\textwidth}
  \centering
  \includegraphics[width=3.15in]{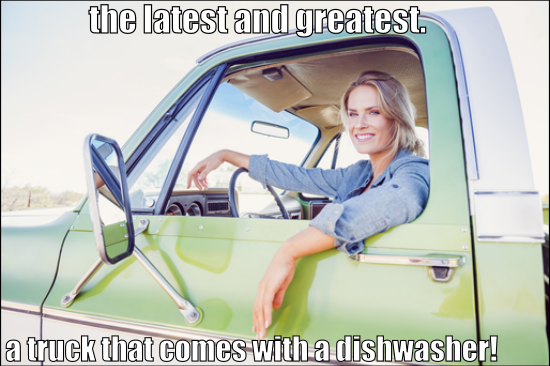}
\endminipage\hfill
\minipage{0.5\textwidth}
  \centering
  \includegraphics[width=3.15in]{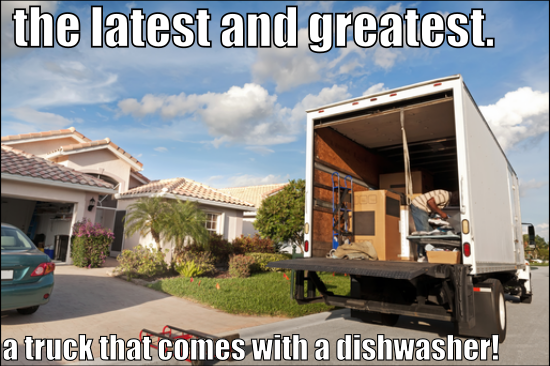}
\endminipage
\caption[An example of a multimodal hateful meme with a benign image confounder counterpart.]{An example of a multimodal hateful meme with a benign image confounder counterpart. MARMOT correctly classifies the meme on the left as hateful and the meme on the right as non-hateful. \copyright Getty Images.}
\label{fig:benign_image1}
\end{figure}

We look at another example, shown in Figure \ref{fig:benign_image1} where the text is the same, but the images are different. MARMOT, again, correctly classifies the meme on the left as hateful and correctly classifies the meme on the right as non-hateful.\footnote{Specifically, the probability of each meme being hateful, using MARMOT, is 0.987 and 0.225, respectively.} 

\begin{figure}[!htbp]
\minipage{0.5\textwidth}
  \centering
  \includegraphics[width=3in]{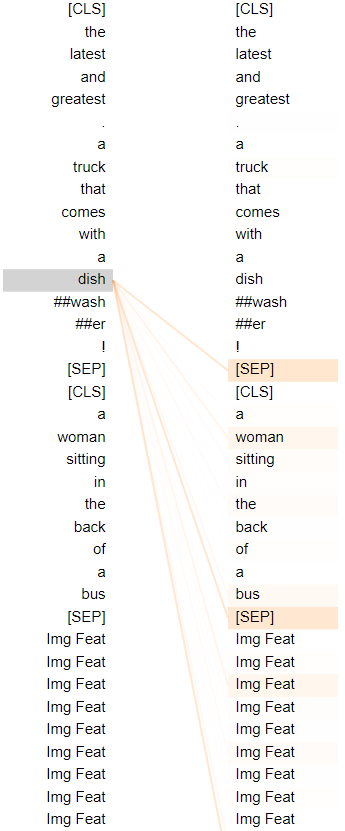}
\endminipage\hfill
\minipage{0.5\textwidth}
  \centering
  \includegraphics[width=3in]{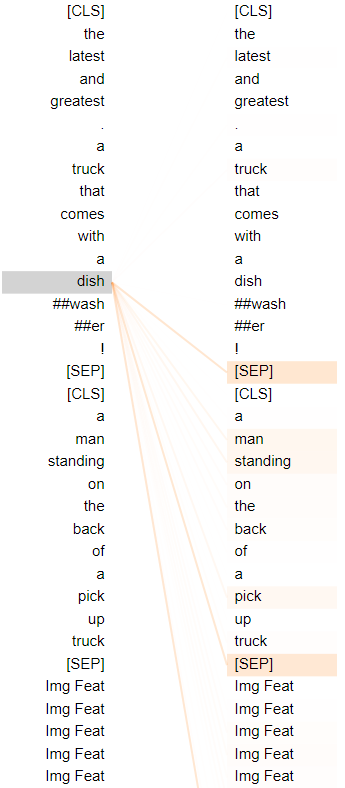}
\endminipage
\caption[A visualization of the attention weights from the 12th attention head of the 10th layer of the BERT encoder from the last step of MARMOT for the two memes from Figure \ref{fig:benign_image1}.]{A visualization of the attention weights from the 12th attention head of the 10th layer of the BERT encoder from the last step of MARMOT for the two memes from Figure \ref{fig:benign_image1}. The visualizations on the left and right correspond with the hateful and non-hateful memes, respectively, from Figure \ref{fig:benign_image1}. The line weight indicates the attention weight. Visualizations were created using the package described in \citet{vig2019transformervis}. The token ``Img Feat'' indicates an inputted image feature, which does not translate to a word token.} 
\label{fig:benign_image1_attentionweights}
\end{figure}

To take a closer look at what the model learns between the image and the text, we again create visualizations of the attention weights of the BERT encoder used in the last stage of MARMOT. Figure \ref{fig:benign_image1_attentionweights} shows the attention weights of the 12th attention head of the 10th layer in the BERT encoder. Again, the line weight reflects the attention weight between the attending tokens and the attended tokens.  In the hateful example, the token ``dish'' shares a high attention weight with the token ``woman'' from the image caption. In the non-hateful example, the token ``dish'' shares a high attention weight with the token ``man'' from the image caption. Again, the attention weights suggest that MARMOT learns different relationships among the images and text between the non-hateful and hateful examples, even when the text is the same. 

\section{Conclusion and Future Directions}
Labeling is usually required to identify posts of interest for most content analyses of social media posts. Human coders usually consider image and text if both are available. However, automated machine approaches are almost always unimodal, focusing exclusively on either the text or images. This can create potential biases in downstream analyses of the machine classified data. Ways of representing text or image as quantitative features are well-known in the computer science literature, but multimodal representations---joint representations of image and text---are still a budding field in computer science. Many state-of-the-art multimodal models require that every observation have both image and text and require extensive pretraining on image annotation datasets. The computational costs and the reality that many datasets of interest for social scientists contain observations with missing modalities render these state-of-the-art multimodal models essentially unusable. 

We present a new method that calculates joint image-text representations called multimodal representations using modality translation, or MARMOT, to solve both issues. It eschews pretraining, meaning that training and inference can be completed with minimal computational resources. It also leverages off-the-shelf pretrained models: good performance results can be achieved on relatively small datasets. Lastly, it can calculate representations for observations missing modalities. 

Specifically, we first note that the pretraining over image annotation datasets used in models such as VisualBERT \citep{li2019visualbert} is done to adapt the underlying BERT model to accept both image and text features, rather than only the text features it was initially pretrained on, and to relate image features with text features. Additional pretraining with the data of interest is done to adapt further the model to the target domain.  

To capture the same spirit of this process without undergoing the computationally expensive pretraining procedure, we develop a process called modality translation. First, we generate image captions directly using a pretrained image captioner. To learn patterns between the image and its image caption, we use a transformer decoder initialized with pretrained BERT weights. The image captions are inputted into the BERT decoder, and the image features, derived from a pretrained image model, are inputted at the encoder-decoder attention layer. This process calculates what we call the translated image. 

After obtaining the image caption and the translated image through modality translation, we jointly input the text of the observation, the text of the image caption, and the translated image as one sequence into the transformer encoder initialized with pretrained BERT weights. The three parts of the sequence are differentiated using different token type embeddings. The joint representation is either the outputted vector corresponding to the \texttt{[CLS]} token or the average across all vectors of the outputted sequence. 

We apply MARMOT to two classification tasks: classifying tweets of election incidents during the 2016 U.S. general election \citep{Mebane.Wu.Woods.Klaver.Pineda.Miller2018} and identifying hateful memes \citep{kiela2020hateful}. In the election incidents dataset, all observations have text, but only some have images. MARMOT outperforms the ensemble classifier found in \citet{Mebane.Wu.Woods.Klaver.Pineda.Miller2018} in 19 of 20 categories in multilabel classifications of tweets (and equals the performance in the last category). However, with the election incidents dataset, we cannot use other state-of-the-art multimodal classifiers because some observations do not contain an image. We turn to the Hateful Memes dataset, where all observations have both text and image. MARMOT improves upon the benchmark state-of-the-art multimodal models in terms of accuracy and area under the receiver operating characteristic curve (AUC), even outperforming pretrained multimodal classifiers. MARMOT improves the best result set by VisualBERT in terms of accuracy from 0.6473 to 0.6760 and in terms of AUC from 0.7141 to 0.7530. Using MARMOT in a deep ensemble further improves accuracy to 0.6920. Qualitatively looking at examples from the test set, MARMOT can correctly classify multimodal hate memes---the memes where the text alone and the image alone are not hateful but combined are hateful. Visualizations of the attention weights used in the BERT encoder, the final stage of MARMOT, further suggest that the architecture is learning important associations between the text and image when making predictions. Thus, MARMOT does not trade off performance on classification problems to accommodate missing modalities. It performs just as well if not better than benchmark state-of-the-art multimodal models while having this key property that makes it useful for political science, communications, and social media research. 

There are still many future directions for this project. Numerous methodological details still require experimentation, such as using newer pretrained language models, using image regions derived from a pretrained image segmentation model such as Faster R-CNN \citep{ren2015faster} instead of image features from ResNet, using different training strategies, and using MARMOT in conjunction with other pretrained vision-and-language models if all observations have both image and text. We also aim to extend MARMOT to non-social media applications, such as multimodal elements in newspapers and other forms of media. 

We have also not used MARMOT representations in any substantive applications, such as using the predictions in a regression framework. In a regression framework, MARMOT can be potentially used in two ways. First, the predictions from using MARMOT representations can be used as a covariate. Second, the predictions from using MARMOT representations can be used as an outcome variable. Using the predictions using MARMOT representations as either the outcome variable or covariate without any adjustments can lead to bias or uncontrolled variance. \citet{fong_tyler_2020} discuss how to adjust machine predictions when used as covariates, and \citet{Wang30266} discuss how to adjust machine predictions when used as the outcome variable. The two frameworks apply to machine prediction pipelines generally, meaning such frameworks can be jointly used with MARMOT to make its predictions useful in substantive applications.  

\newpage
\begin{singlespace}
\bibliographystyle{newapa_full}
\bibliography{refs.bib}
\end{singlespace}

\newpage
\section{Supplemental Information}
\label{sec:si_p2}

\subsection{Feedforward Neural Networks}
\label{sec:ffnn_si_p2}
Define a single-layer feedforward neural network as $s = XW_1$, where $X$ is the design matrix of dimension $N \times d$ and $W_1$ is $d \times C$, where $N$ is the total number of observations, $d$ is the total number of input features, and $C$ is either 1 when using neural networks for regression problems or $C$ is the total number of classes that an observation can belong to. Including a column of 1's in the design matrix adds a bias term. This produces a score matrix $s$ of dimension $N \times C$. In a classification problem, each row contains the raw scores of an observation belonging to each class $c$. The argmax of each row is the predicted class. 

We can plausibly give this network more representational power by using a second weight function, $W_2$, of dimension $k \times C$, and redefining $W_1$ as a $d \times k$ matrix, where $k$ is a dimension that we can choose and is often referred to as the number of hidden neurons. This is known as a two-layer neural network. Notice that we cannot simply define the two-layer neural network as $s = (XW_1)W_2$, because this simply collapses back into a single-layer neural network: $s = (XW_1)W_2 = XW$ where $W = W_1 W_2$ and has dimensions $d \times C$. Instead, we introduce a nonlinear function, $f$, called the activation function. This nonlinear function is applied pointwise. We can then properly define the two-layer neural network as $s = f(XW_1)W_2$. Popular options for $f$ include the inverse $tan$ function, the sigmoid function, $\sigma(x) = \frac{1}{1+e^{-x}}$, and the ReLU function, $ReLU(x) = max(0,x)$. Because of its simplicity and its empirically-determined robustness, the ReLU function is the most popular activation function. 

A three-layer neural network would be similarly defined: redefine $W_1$ as a $d \times k_1$ matrix, redefine $W_2$ as a $k_1 \times k_2$ matrix, and define $W_3$ as a $k_2 \times C$ matrix; here, $k_1$ and $k_2$ are the number of hidden neurons. Then, $s = f(f(XW_1)W_2)W_3$. $s$ is ultimately still an $N \times C$ matrix, with prediction carried out in the same way as the single-layer neural network. 

The ultimate goal of the model is to have weights that minimize prediction errors. To do this, we need a measure of how well the model predicts and a way to update the weights of the network such that it reduces prediction errors. A loss function quantifies how well the model performs by comparing predictions with ground truth data. Popular loss functions include the cross-entropy loss function for classification problems and the mean squared error for regression problems. We can then use backpropagation \citep{backpropagation} to calculate the gradient of the loss function: we calculate the gradient of the loss function with respect to each weight one layer at a time, iterating from the last layer to the first layer. Weights are then updated using a gradient descent step.  

\subsection{Residual Connections}
Theoretically, a deeper neural network should be able to perform just as well as a shallower neural network. A deeper network can imitate a shallower network by copying over the layers of the shallower network and setting the additional layers to the identity mapping. But \citet{he2015deep} observe that deeper networks often perform worse than shallower networks. They argue that as a neural network increases in layers it is harder for information from shallower layers to propagate to deeper layers. To solve what they call the degradation problem, they propose a very simple solution: after a set of layers $F(\mathbf{x})$, we sum the initial input $\mathbf{x}$ and the learned mapping $F(\mathbf{x})$. In other words, this set of layers outputs $F(\mathbf{x}) + \mathbf{x}$ instead of $H(\mathbf{x})$. This allows information from shallower layers to propagate forward more easily. It is called a residual connection because the layers learns the residual, or information not immediately learned from $\mathbf{x}$ directly. Although simple in concept, residual connections allowed much deeper networks to be trained. 

\subsection{Layer Normalization}
\label{sec:ln}
Gradients with respect to weights in one layer depend on the outputs of the previous layer. This can cause the distribution of the parameters of one layer to shift in a way that affects the quality of learning for deeper layers. A normalization procedure, such as layer normalization, can reduce covariate shift \citep{ba2016layer}. Layer normalization normalizes the inputs across the features. We can recalculate the inputs at each layer as follows: 
\[\mu_i = \frac{1}{K} \sum_{k=1}^{K} x_{ik}\]
\[\sigma_i^2 = \frac{1}{K} \sum_{k=1}^{K} (x_{ik}-\mu_i)^2\]
\[\hat{x_{ik}} = \frac{x_{ik} - \mu_i}{\sqrt{\sigma_i^2 + \varepsilon}}\]
where $x_{ik}$ is the $k$th feature of the $i$th sample and $K$ is the total number of features. The last step is to scale and shift $\hat{x_i}$ by $\gamma$ and $\beta$, respectively, which are learnable parameters: $\text{LN}_{\gamma,\beta}(x_i) = \gamma \hat{x_i} + \beta$. 

\subsection{Additional Details About BERT}
\label{sec:more_details_bert}
There are two variants of BERT. $\text{BERT}_{\text{base}}$ is 12 transformer encoder blocks stacked, while $\text{BERT}_{\text{large}}$ is 24 transformer encoder blocks stacked. The former uses embeddings with 768 dimensions, while the latter uses embeddings with 1,024 dimensions. $\text{BERT}_{\text{base}}$ is 110 million parameters, while $\text{BERT}_{\text{large}}$ is 340 million parameters. BERT takes as input a sequence of WordPiece embeddings \citep{wu2016googles}. 

In MLM, 15\% of words are masked out and BERT must predict what the masked out word is. In NSP, pairs are sentences are inputted. Half of these pairs' second sentence is the sentence that follows the first sentence in the original corpus; the other half contains sentences randomly paired together. BERT must predict if the second sentence actually follows the first sentence in the original corpus. This prediction is made using the corresponding output vector for the \texttt{[CLS]} token.

Popular word embedding methods such as word2vec or GloVe embeddings \citep{pennington2014glove} are context-free, meaning that each word is assigned a fix embedding irrespective of its context. For example, the word2vec embedding for the word ``trump'' would be the same in the sentences ``I support Donald Trump'' and ``She played the trump card,'' despite the fact that ``trump'' has different meanings in each sentence. In other words, word2vec or GloVe word embeddings cannot map multiple vectors for polysemous words in a self-supervised fashion. BERT embeddings, on the other hand, are context-dependent: the BERT embedding for a given word is different for each context. The self-attention mechanism within the transformer encoder outputs a word embedding that, in essence, is a weighted average of all the other word embeddings of words in the document. The word embedding for ``trump'' in each sentence would be different.

\subsection{Definition of Evaluation Metrics}
\label{sec:evalmetrics}

\subsubsection{Accuracy, Precision, Recall, and F1 Score} 
In binary classification problems, predictions are either for the positive or negative class. We can evaluate our predictions against the ground truth using the terms true positives (TP), true negatives (TN), false positives (FP), and false negatives (FN). False positives are known as Type I errors, while false negatives are known as Type II errors. With the predictions sorted into one of these four categories, we can calculate accuracy, precision, recall, and the F1 score. The metric most important to consider varies from problem to problem, although the F1 score is often used as an overall metric of performance. Precision, recall, and F1 are defined individually over each of the two categories. Macro and micro F1 are defined by combining the F1 metrics across the two classes. Accuracy is defined across the two classes. 

\[\text{Accuracy} = \frac{TP+TN}{TP+TN+FP+FN}\]

\[\text{Precision}_0 = \frac{TN}{TN+FN}\]

\[\text{Precision}_1 = \frac{TP}{TP+FP}\]

\[\text{Recall}_0 = \frac{TN}{TN+FP}\]

\[\text{Recall}_1 = \frac{TP}{TP+FN}\]

\[\text{F1}_0 = 2 \cdot \frac{\text{Precision}_0 \text{Recall}_0}{\text{Precision}_0 + \text{Recall}_0} \]

\[\text{F1}_1 = 2 \cdot \frac{\text{Precision}_1 \text{Recall}_1}{\text{Precision}_1 + \text{Recall}_1} \]

\noindent Denoting $N = N_0 + N_1$, where $N_0$ is the number of observations that belong to the negative class and $N_1$ is the number of observations that belong to the positive class, 

\[\text{Macro F1} = \frac{\text{F1}_0 + \text{F1}_1}{2}\]

\[\text{Micro F1} = \frac{N_0}{N}\text{F1}_0 + \frac{N_1}{N}\text{F1}_1\]

\subsubsection{Area Under the Receiver Operating Characteristic Curve (AUC)} 
\begin{figure}[!ht]
    \centering
    \includegraphics[scale=0.7]{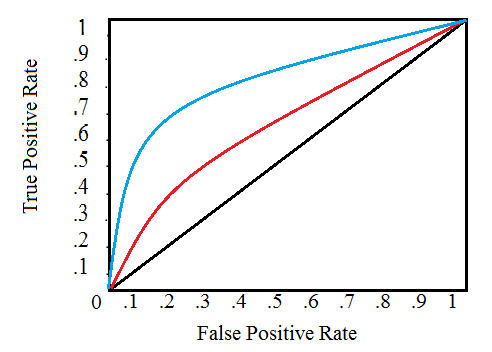}
    \caption[Example of an ROC curve.]{Example of an ROC curve. Figure taken from \url{https://www.statisticshowto.com/receiver-operating-characteristic-roc-curve/}.} 
    \label{fig:roc_example}
\end{figure}

The receiver operating characteristic curve (ROC curve) plots the true positive rate (or recall) against the false positive rate at various threshold settings. The true positive rate, or TPR, is defined as $\frac{TP}{TP+FN}$ while the false positive rate, or FPR, is defined as $\frac{FP}{FN+TP}$. When we make a classification, notice that the model does not simply return a 1 or 0; instead, it returns a probability that an observation belongs to the positive class. For example, we usually classify any inputted observation that returns a probability greater than or equal to 0.5 as the positive class. But this threshold is arbitrary. If we increase the threshold, the true positive rate would drop but so would the false positive rate. If we decreased the threshold, the true positive rate would rise but so would the false positive rate. See Figure \ref{fig:roc_example} for an example of an ROC curve. An observation under the diagonal line signals worse-than-random; an observation on the line signals random guessing; an observation at (0,1) indicates a perfect classifier---it has a perfect TPR and the FPR is 0.  

A way of assessing the ROC curve is to measure the area under the curve, or AUC. A perfect AUC would be 1, corresponding to the 90 degree curve that consists of a line segment from (0,0) to (0,1) and a line segment from (0,1) to (1,1), indicating a perfect classifier. More interestingly, the AUC can be shown to be equivalent to the Mann-Whitney U statistic \citep{elements_of_stat_learning}. The AUC is equal to the probability that a classifier will rank a randomly chosen positive instance higher than a randomly chosen negative instance. Thus, the AUC can be interpreted as how well the classifier discriminates between the two classes. For this reason, it is often the preferred metric over other metrics, like accuracy or F1. Its major drawback is interpretability. 

\subsection{Additional Details about VirTex}
\label{sec:virtex_details_si_paper2}
\begin{figure}[!htbp]
    \centering
    \includegraphics[scale=0.32]{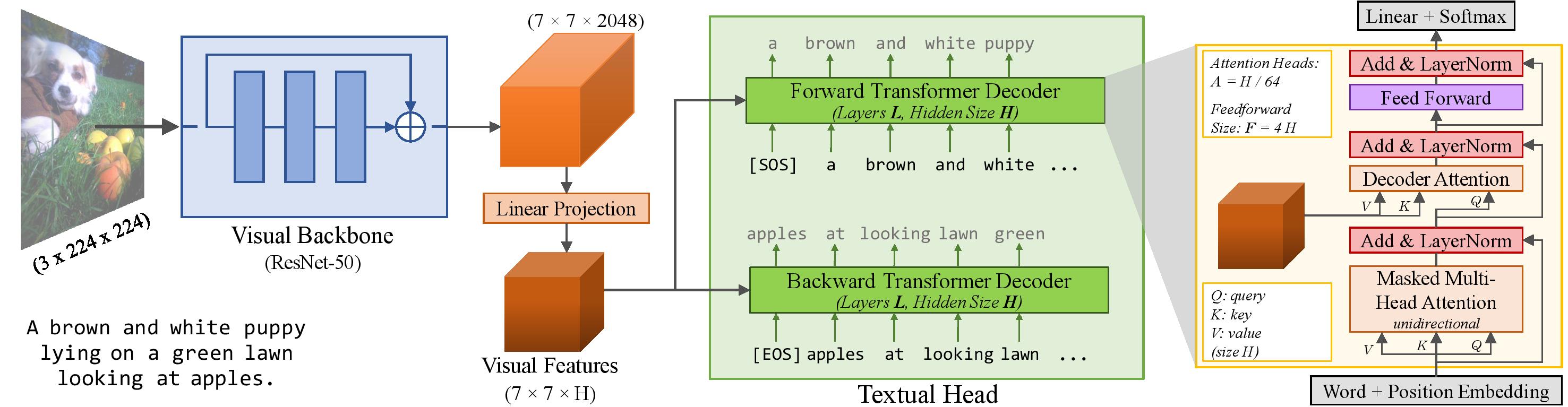}
    \caption[The VirTex pretraining setup, as described in \citet{desai2021virtex}.]{The VirTex pretraining setup, as described in \citet{desai2021virtex}. This figure is taken directly from \citet{desai2021virtex}.}
    \label{fig:virtex_pretraining_setup}
\end{figure}

\citet{desai2021virtex} propose pretraining a deep convolutional neural network (ConvNet) using images and image captions. This differs from the typical approach of pretraining ConvNets, which typically uses a large, labeled image dataset such as ImageNet \citep{imagenet}. Their goal is to learn high quality image representations while using much fewer images. To do this, they jointly train a ConvNet and a transformer \citep{vaswani.2017} using image-caption pairs. They use ResNet-50 as the ConvNet. The visual backbone extracts image features, and then the transformer predicts the caption. The model is then trained end-to-end from scratch. After this pretraining process, the ConvNet can be used for downstream visual recognition tasks. See Figure \ref{fig:virtex_pretraining_setup} for an overview of the pretraining setup. 

\subsection{Additional Details about Self-Critical Sequence Training}
\label{sec:scst}
We use self-critical sequence training \citep{rennie2016selfcritical} to calculate the image captions. Deep generative models typically use a technique known as teacher-forcing, which maximizes the likelihood of the next ground-truth word given the previous ground-truth words. This creates a mismatch between training and testing---during testing, the previous generated words are used instead of the previous ground-truth labels. \citet{rennie2016selfcritical} approach the image captioning problem with a reinforcement learning framework. The recurrent models, the LSTMs, are the ``agents'' that interacts with an external ``environment'' consisting of words and image features. The parameters of the network define a policy $p_\theta$. When the end-of-sentence token is generated, or EOS, the agent is given a ``reward'' that is computed by evaluating generated caption with the ground-truth caption using some kind of a metric, such as CIDEr. This model is pretrained on Microsoft COCO \citep{lin2014microsoft}. We do not further train this model; instead, we simply put the pretrained model in inference mode and generated a caption (or multiple captions) per image. 

\subsection{Details on Training MARMOT} 
To train MARMOT, the following hyperparameter choices need to be made: 
\begin{singlespace}
\begin{itemize}
    \item Using either $\text{BERT}_{\text{base}}$ or $\text{BERT}_{\text{large}}$; refer to the Supplemental Information for more information about $\text{BERT}_{\text{base}}$ and $\text{BERT}_{\text{large}}$. The primary difference is the number of encoder layers used to pretrain each model.
    \item The number of training epochs
    \item The learning rate
    \item The learning rate schedule
    \item The batch size
    \item The number of captions to use per image
    \item Whether to freeze certain parts of the architecture 
    \item The details of the fully-connected classifier
\end{itemize}
\end{singlespace}
In our applications, we use $\text{BERT}_{\text{base}}$, primarily because of computational constraints. We select the training epochs from $\{3,4\}$, the learning rate from $\{2 \times 10^{-5}, 3 \times 10^{-5}, 5 \times 10^{-5}\}$, and the batch size from $\{16, 32\}$. The limited ranges of hyperparameters make grid searches feasible. These are the same values suggested in \citet{devlin.chang.lee.toutanova.2018}. We use the cosine learning rate schedule \citep{huang2017snapshot}. We find that using 3 image captions generally work well, with more captions typically worsening performance. Using the weighted Adam optimizer \citep{kingma2014adam}, as suggested in \citet{devlin.chang.lee.toutanova.2018}, generally works well. 

Whether we freeze certain parts of the architecture depends on the application and grid search feasibility. \citet{wang2019makes} notes that part of the difficulty of training multimodal models is that the image and text features are learned by the model at different rates. One strategy to alleviate this issue is to freeze and unfreeze certain parts of the model at different stages of training \citep{kiela2019supervised}. When a part of the model is frozen, the parameters are not updated after the backwards pass. In the first few iterations, both the transformer decoder of modality translation and the pretrained language model are frozen, allowing the model to only have the ability to update the weights of the $1 \times 1$ convolution over the image features and the pretrained image network. Then, the transformer decoder of modality translation is unfrozen but the pretrained language model remains frozen. Lastly, the pretrained model is unfrozen and the model is trained end-to-end. 

Finetuning is also quite sensitive to the learning rate schedule, which is how the learning rate changes as training progresses. We design a three-stage learning rate scheduler when freezing certain parts of the architecture. The first 10\% of the total training iterations are dedicated to a warmup period, where the learning rate rises from 0 to the initial set learning rate in a linear fashion. Then for the epochs where the transformer decoder of modality translation or pretrained language model are frozen the learning rate is fixed at the initial set learning rate. When finetuning of the entire model occurs after the pretrained language model is unfrozen, the learning rate decreases following the values of a cosine function between the initial set learning rate to zero. Freezing the transformer decoder of modality translation for 2 epochs and the pretrained language model for 4 epochs generally worked well, and experimentation showed that more epochs where either parts were frozen did not yield improvements in performance. 

The model is implemented in Python using PyTorch and HuggingFace's \texttt{transformers} library \citep{wolf2019huggingfaces}. 

\subsection{Application 1: Definitions of Subcategories}
\label{sec:category_and_adjective_definitions}
The sub-bullet points indicate the subcategories that belong to a given category. More detailed definitions for each category can be found in \citet{Mebane.Wu.Woods.Klaver.Pineda.Miller2018}.

\begin{singlespace}
\begin{itemize}
    \item Line length, waiting time, polling place overcrowding
    \begin{itemize}
        \item There was no crowd or line at the polling place
        \item There was a small crowd, short line, or wait 
        \item The polling place was crowded or there was a long line or wait (20 minutes or longer)
    \end{itemize}
    \item Polling place event
    \begin{itemize}
        \item The polling place did not function as expected or information is incorrect 
        \item The tweet describes the polling place without noting whether it or an aspect functioned correctly or incorrectly
        \item The polling place did function correctly or information is correct
    \end{itemize}
    \item Electoral system 
    \begin{itemize}
        \item The electoral system did not function appropriately 
        \item The tweet makes a neutral statement about the electoral system without an indication of if it functioned appropriately 
    \end{itemize}
    \item Absentee, mail-in, or provisional ballot issue 
    \begin{itemize}
        \item The absentee, mail-in, or provisional ballot system did not function appropriately 
        \item The tweet makes a neutral observation or statement about the absentee, mail-in, or provisional ballot system without noting it having functioned correctly or incorrectly
        \item The absentee, mail-in, or provisional ballot system functioned properly
    \end{itemize}
    \item Registration
    \begin{itemize}
        \item The tweet indicates that an individual was not able to register to vote
        \item The tweet makes a neutral observation about the voter registration process without noting if the individual in question registered or not
        \item The tweet notes that the individual was able to vote 
    \end{itemize}
\end{itemize}
\end{singlespace}

\subsection{Application 1: Hyperparameters}
\label{sec:election_hp}
To create a validation dataset, we randomly held out 20\% of the training set for each category and subcategory. We used a grid search to find the learning rate and the number of training epochs that optimized the F1 score over the class. We did not grid search over the batch size because of computational constaints. Table \ref{tab:election_hp} details the hyperparameters for the categories and subcategories. 

\begin{table}[!htbp]
\centering
\begin{tabular}{l|ccc}
\textbf{Category}              & \multicolumn{1}{l}{\textbf{Batch Size}} & \multicolumn{1}{l}{\textbf{Learning Rate}} & \multicolumn{1}{l}{\textbf{Epochs}} \\ \hline
\textbf{Not an Incident}       & 16                                      & $5 \times 10^{-5}$                         & 4                                   \\
\textbf{Line Length}           & 16                                      & $2 \times 10^{-5}$                         & 4                                   \\
(a) No crowd                   & 16                                      & $3 \times 10^{-5}$                         & 4                                   \\
(b) Small crowd                & 16                                      & $3 \times 10^{-5}$                         & 4                                   \\ 
(c) Large crowd                & 16                                      & $5 \times 10^{-5}$                         & 4                                   \\
\textbf{Polling Place Event}   & 16                                      & $3 \times 10^{-5}$                         & 4                                   \\
(a) Did not function as expected & 16                                    & $2 \times 10^{-5}$                         & 4                                   \\
(b) Neutral observation        & 16                                      & $5 \times 10^{-5}$                         & 4                                   \\
(c) Functioned properly        & 16                                      & $2 \times 10^{-5}$                         & 3                                   \\
\textbf{Electoral System}      & 16                                      & $3 \times 10^{-5}$                         & 4                                   \\
(a) Did not function properly  & 16                                      & $3 \times 10^{-5}$                         & 3                                   \\
(b) No comment on function     & 16                                      & $2 \times 10^{-5}$                         & 4                                   \\
\textbf{Absentee or Early Voting Issue} & 16                             & $5 \times 10^{-5}$                         & 4                                   \\
(a) Did not function properly  & 16                                      & $5 \times 10^{-5}$                         & 4                                   \\
(b) Neutral observation        & 16                                      & $5 \times 10^{-5}$                         & 4                                   \\
(c) Functioned properly        & 16                                      & $5 \times 10^{-5}$                         & 4                                   \\
\textbf{Registration}          & 16                                      & $2 \times 10^{-5}$                         & 4                                   \\
(a) Not able to register       & 16                                      & $2 \times 10^{-5}$                         & 4                                   \\
(b) Neutral observation        & 16                                      & $5 \times 10^{-5}$                         & 4                                   \\
(c) Able to register           & 16                                      & $3 \times 10^{-5}$                         & 4
\end{tabular}
\caption{The selected hyperparameters for each category and subcategory for the tweets about election incidents during the 2016 U.S. general election.} 
\label{tab:election_hp}
\end{table}

We used $\text{BERT}_{\text{base}}$. We did not freeze any parts of MARMOT. We used weighted Adam \citep{kingma2014adam} with a cosine learning schedule. We did not use gradient clipping. The $\varepsilon$, $\beta_1$, and $\beta_2$ for Adam are set to $1 \times 10^{-8}$, $0.9$, and $0.98$ respectively, following the parameters set by \citet{kiela2020hateful}. 

\subsection{Application 2: Hyperparameters}
\label{sec:hm_hp}
Hyperparameters were selected using the dev set provided by the Hateful Memes dataset \citep{kiela2020hateful}. We used a grid search to find the batch size, learning rate, and number of training epochs that optimized area under the receiver operating characteristic curve (AUC) over the dev set. The hyperparameters selected were: batch size of 32, learning rate of $5 \times 10^{-5}$, and 8 training epochs. We used $\text{BERT}_{\text{base}}$. 

The parameters of both the transformer decoder of modality translation and the pretrained language model were frozen for 2 epochs, and the parameters of just the pretrained language model were frozen for 2 more epochs after. After the 4th epoch, the entire model is finetuned end to end. We used weighted Adam \citep{kingma2014adam} with the learning rate schedule detailed in the Supplemental Information. We did not use gradient clipping. The $\varepsilon$, $\beta_1$, and $\beta_2$ for Adam are set to $1 \times 10^{-8}$, $0.9$, and $0.98$ respectively, following the parameters set by \citet{kiela2020hateful}.

\subsection{Application 2: Accuracy Learning Curve During Training over the Hateful Memes Dataset}
\label{sec:si_hatefulmemes_acc_curve}

\begin{figure}[!htbp]
    \centering
    \includegraphics[width=5in]{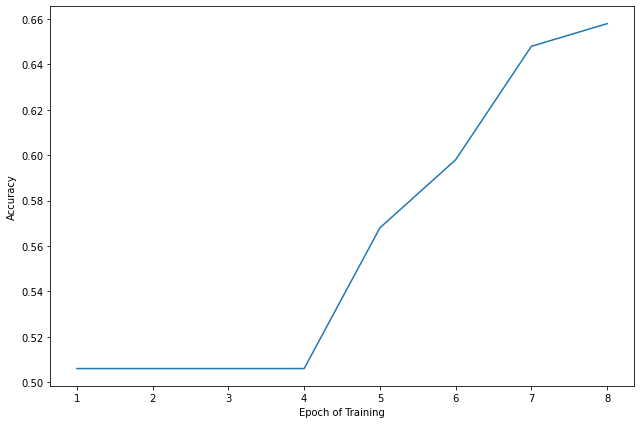}
    \caption{Accuracy learning curve of the validation set during training over the Hateful Memes dataset.}
    \label{fig:accuracy_curve_hm}
\end{figure}

To assess whether MARMOT overfits the training data, we plot the accuracy learning curve of the validation set during training. Figure \ref{fig:accuracy_curve_hm} shows that the curve monotonically increases during training. Because the parameters of both the transformer decoder of modality translation and the pretrained language model were frozen for 2 epochs and the parameters of just the pretrained language model was frozen for 2 more epochs after, the accuracy does not increase in the first four epochs of training. 

\subsection{Application 2: Results Over the Validation Set of the Hateful Memes Dataset} 
\label{sec:si_hatefulmemes_validation}
Table \ref{tab:hatefulmemes_validation} contains the accuracy and area under the receiver operating characteristic curve (AUC) performance metrics across the 11 baseline models and MARMOT over the validation dataset. Results for the 11 baseline models come from \citet{kiela2020hateful}. It is important to note that hyperparameters were optimized using the validation dataset. The results are largely in line with the results over the test set. 

\begin{table}[!htbp]
\centering
\begin{tabular}{l|cc}
    \textbf{Model}  & \textbf{Accuracy} & \textbf{AUC}    \\ \hline
    Image - Grid    & 0.5273            & 0.5879          \\
    Image - Region  & 0.5266            & 0.5798          \\
    Text BERT       & 0.5826            & 0.6465          \\
    Late Fusion     & 0.6153            & 0.6597          \\
    Concat BERT     & 0.5860            & 0.6525          \\
    MMBT - Grid     & 0.5820            & 0.6857          \\
    MMBT - Region   & 0.5873            & 0.7173          \\
    ViLBERT         & 0.6220            & 0.7113          \\
    VisualBERT      & 0.6210            & 0.7060          \\
    ViLBERT CC      & 0.6140            & 0.7060          \\
    VisualBERT COCO & 0.6506            & 0.7397          \\
    MARMOT          & 0.6580            & 0.7587     
    \end{tabular}
    \caption{Accuracy and area under the receiver operating characteristic curve (AUC) performance metrics across the 11 baseline models and MARMOT over the validation set of the Hateful Memes dataset.} 
    \label{tab:hatefulmemes_validation}
\end{table}

\end{document}